# Generative AI in the Construction Industry: A State-of-the-art Analysis

Ridwan Taiwo[1*], Idris Temitope Bello[1,2], Sulemana Fatoama Abdulai[1], Abdul-Mugis Yussif[1], Babatunde Abiodun Salami[3], Abdullahi Saka[4], Tarek Zayed[1*]

[1]Department of Building and Real Estate, The Hong Kong Polytechnic University, Hung Hom, Kowloon, Hong Kong, China.

[2]Centre for Advances in Reliability and Safety (CAiRS), Hong Kong Science Park12/F, Building 19W, Pak Shek Kok, NT, Hong Kong, China

[3]Cardiff School of Management, Cardiff Metropolitan University, Llandaff Campus, Cardiff CF5 2YB, United Kingdom.

[4]School of Built Environment, Engineering and Computing, Leeds Beckett University, UK

Corresponding authors: Ridwan Taiwo (ridwan-a.taiwo@connect.polyu.hk) and Tarek Zayed (tarek.zayed@polyu.edu.hk)

## Graphical Abstract

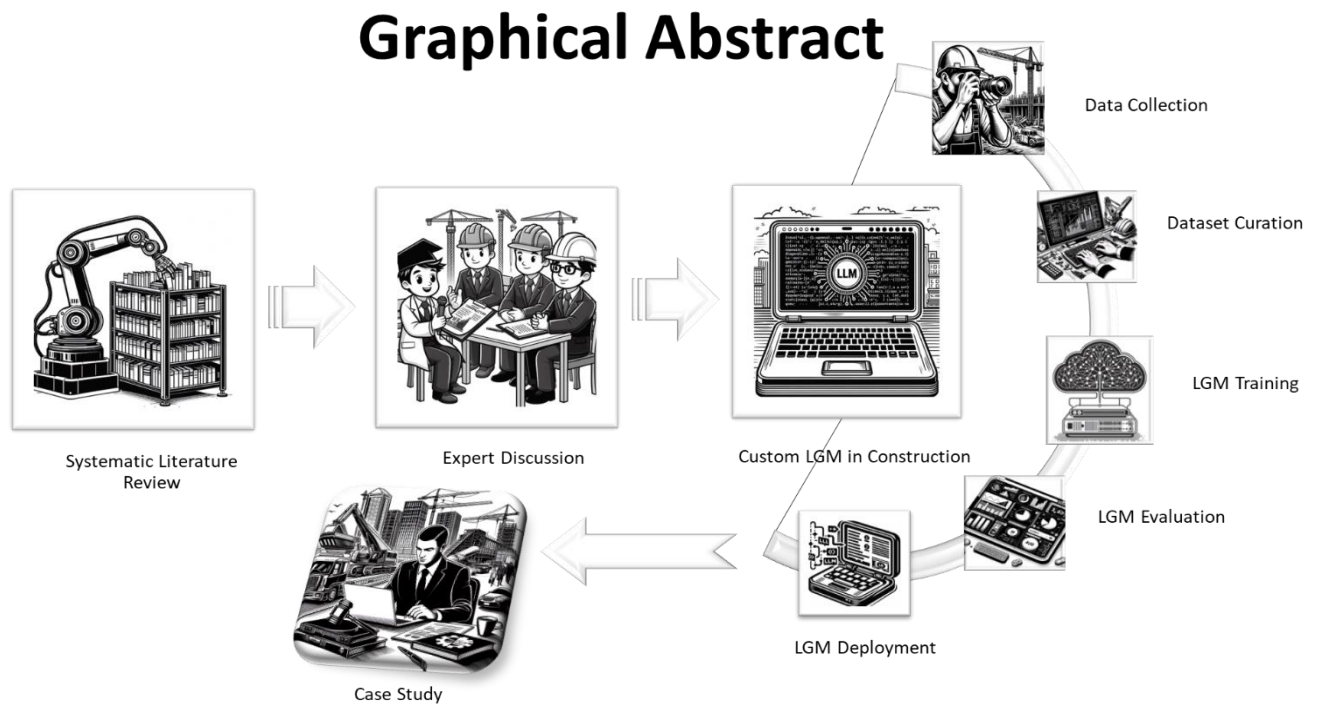

# Abstract

The construction industry is a vital sector of the global economy, but it faces many productivity challenges in various processes, such as design, planning, procurement, inspection, and maintenance. Generative artificial intelligence (AI), which can create novel and realistic data or content, such as text, image, video, or code, based on some input or prior knowledge, offers innovative and disruptive solutions to address these challenges. However, there is a gap in the literature on the current state, opportunities, and challenges of generative AI in the construction industry. This study aims to fill this gap by providing a state-of-the-art analysis of generative AI in construction, with three objectives: (1) to review and categorize the existing and emerging generative AI opportunities and challenges in the construction industry; (2) to propose a framework for construction firms to build customized generative AI solutions using their own data, comprising steps such as data collection, dataset curation, training custom large language model (LLM), model evaluation, and deployment; and (3) to demonstrate the framework via a case study of developing a generative model for querying contract documents. The results show that retrieval augmented generation (RAG) improves the baseline LLM by 5.2, 9.4, and 4.8% in terms of quality, relevance, and reproducibility. This study provides academics and construction professionals with a comprehensive analysis and practical framework to guide the adoption of generative AI techniques to enhance productivity, quality, safety, and sustainability across the construction industry.

# 1. Introduction

The construction industry is one of the most critical sectors of the global economy, accounting for approximately $10 trillion or 13% of global GDP in 2019 [1–3]. It also employs over 220 million workers globally and is often a significant driver of employment [4]. The construction industry encompasses various sub-sectors, such as civil engineering, infrastructure, residential, commercial, and industrial building construction [1–3]. Among these, building construction is the largest and most diverse sub-sector, accounting for about 40% of the global construction output and 50% of the worldwide construction employment [5].

Building construction is a complex and dynamic process that involves multiple stakeholders, such as owners, architects, engineers, contractors, subcontractors, suppliers, and regulators [6,7]. The process requires coordinating and integrating various activities, such as design, planning, scheduling, procurement, fabrication, installation, inspection, and maintenance [8,9]. The process also generates and consumes vast data, such as drawings, specifications, contracts, reports, invoices, and photos[10]. Building construction's quality, efficiency, and sustainability depend primarily on how well these activities and data are managed and utilized.

However, the construction industry faces many challenges that hinder its performance and productivity. These challenges include the design, construction, procurement and supply chain, fabrication and installation, and inspection and maintenance of buildings, as shown in **Figure 1** [11–13]. Each of these aspects involves complex and dynamic processes that require coordinating and integrating various resources, disciplines, and stakeholders and considering and managing various constraints, uncertainties, and changes. These challenges pose significant difficulties and risks for the construction industry, resulting in low productivity, high cost, long delays, poor quality, and high environmental impact. According to a report by McKinsey, the global construction industry has an average annual productivity growth of only 1%, compared to 2.8% for the total world economy and 3.6% for manufacturing. The report also estimates that the global construction industry could save up to $1.6 trillion per year by improving its productivity to the level of other sectors [1].

Addressing these challenges and improving the performance and productivity of the building construction industry requires innovative and disruptive solutions that can leverage the power of data and technology. One of the most promising and emerging solutions is generative artificial intelligence (AI) [14–16]. Generative AI is a branch of AI that aims to create novel and realistic data or content, such as text, image, video, audio, or code, based on some input or prior knowledge [17]. Generative AI can be seen as the opposite of discriminative AI, which aims to classify or recognize data or content, such as identifying objects in an image or translating text from one language to another. Generative AI can also be seen as a form of creative AI, which aims to produce data or content that is not only realistic but also original, diverse, and expressive [18].

Generative AI also relies on large language models (LLMs), which are neural network models that can generate natural language text based on a given prompt or context. LLMs are trained on massive amounts of text data from various sources, such as books, articles, websites, and social media. They can capture natural language's semantic and syntactic patterns and relationships [19]. Generative AI has been adopted and applied in various disciplines and domains. For instance, in healthcare, generative models have been used to generate synthetic medical images to address data privacy issues and augment datasets to improve disease detection models [20]. Pharmaceutical companies are also exploring generative chemistry to develop new molecular structures for drug discovery [21,22]. Similarly, businesses have leveraged generative design to create new product ideas early in the conceptual process by automating styles and prototyping [23]. In the social sciences, generative AI enhances historical analysis by restoring damaged documents and generating realistic synthetic population data to explore societal issues [24]. Academia is investigating developed personalized study tools and interactive simulated experiences to complement traditional education [25].

Although generative AI applications are still at an infant stage in the construction industry, a few studies exist on their usage for construction-related work [26,27]. As such, [28] reviewed applications of generative AI for developing and enhancing structural designs and how it could help improve accuracy in the design process. [29] presented an overview of the potential applications of Generative Pre-trained Transformer (GPT) models across the lifecycle of a construction project and a case study for material selection. Further, opportunities and a limited number of challenges of adopting generative AI in the construction industry were presented by [30]. Despite the potential and promise of generative AI for the construction industry, there needs to be more systematic and comprehensive literature that reviews and analyzes the current state, opportunities, and challenges of generative AI in this domain. Most existing literature focuses on specific applications or aspects of generative AI (such as GPT models or text-text models) without considering the broader and holistic picture of generative AI in the construction industry. Moreover, there is an urgent need for a more practical and actionable guidance on implementing and deploying generative AI solutions in the construction industry, especially for construction firms that may need more data, expertise, or resources to develop their own generative AI models from scratch.

Therefore, this study aims to provide a state-of-the-art analysis of generative AI in the construction industry, with the following objectives:

- To provide potential opportunities and challenges of applying generative AI in the construction industry by reviewing and categorizing the existing and emerging generative AI applications and use cases.

- To propose an implementation framework for construction firms to build customized generative AI solutions using their data by describing and explaining the key steps and components of the framework.

- To demonstrate the proposed framework via a practical use case of developing a tailored generative model for contract documents.

The rest of the paper is organized as follows: Section 2 briefly overviews the foundational algorithms and large generative models. Section 3 describes the study's methodology, which is explained in four phases. Section 4 presents the literature review results and expert discussion, including the current applications, opportunities, and challenges of generative AI in the construction industry. Section 5 proposes the framework for building custom LLM in the construction industry and explains the main steps and components of the framework. Section 6 demonstrates the framework via a case study of developing a generative model for contract documents and shows the results and outcomes of the case study. Section 7 concludes the paper and provides some directions for future research.

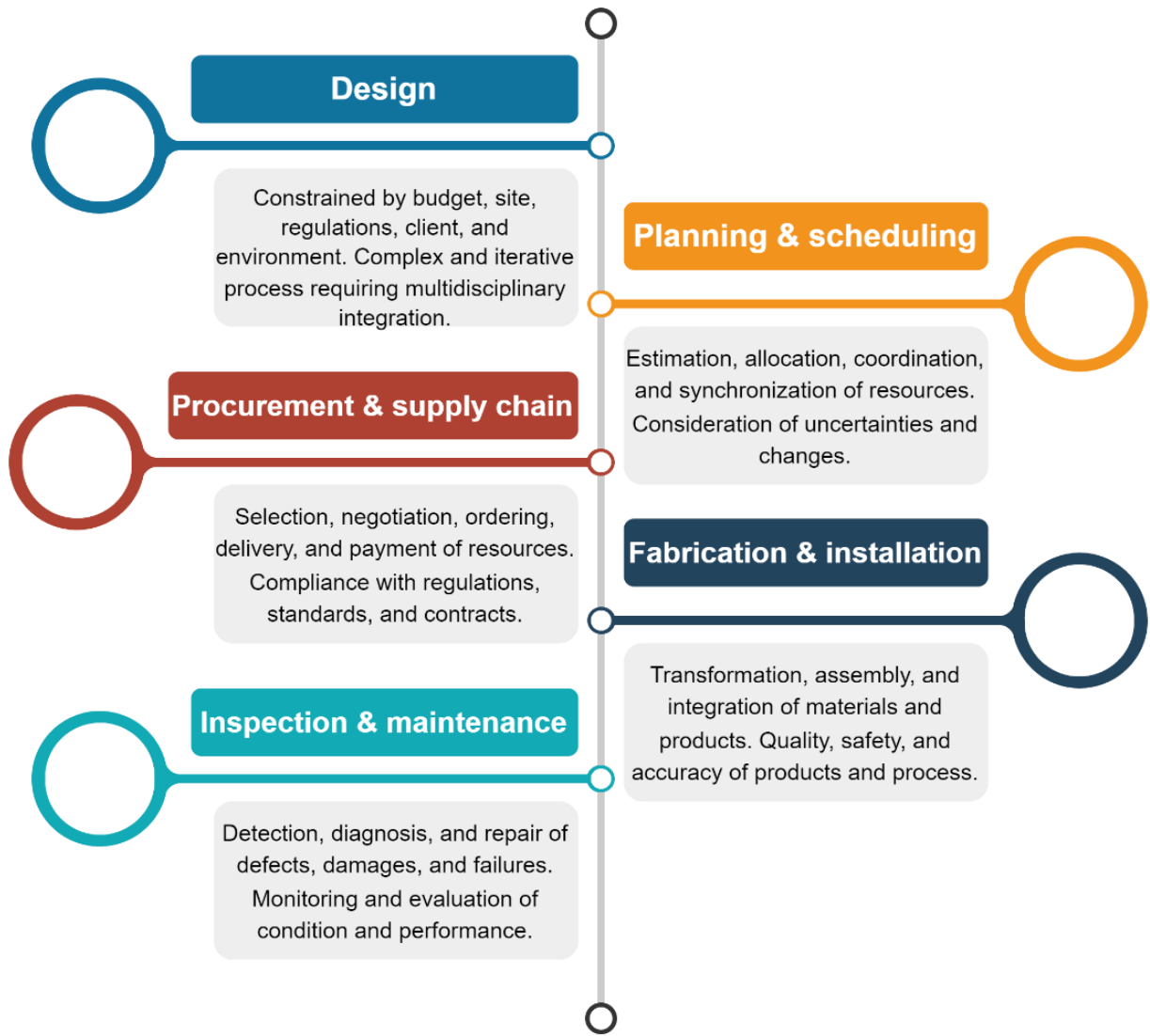

**Figure 1:** Major challenges of the construction industry.

## 2. Generative AI

Generative AI is a branch of AI that aims to create novel and realistic data or content, such as text, image, video, audio, or code, based on some input or prior knowledge [31]. In contrast to discriminative AI [32,33], which aims to categorize or identify existing data, generative AI focuses on creating new, original data or content. For example, while discriminative models may label objects in an image, generative models can produce new images or texts based on specified inputs[24,28].

Generative AI is based on various foundational algorithms, such as generative adversarial networks (GANs), autoregressive models, variational autoencoders (VAEs), transformer models, diffusion models, normalizing flows, and energy-based models. These algorithms use different techniques, such as adversarial learning, probabilistic modeling, attention mechanism, denoising, and density estimation, to learn the underlying distribution or structure of the data or content and generate new samples or variations. Generative AI also relies on large language models (LLMs), which are neural network models that can generate natural language text based on a given prompt or context. LLMs are trained on massive amounts of text data from various sources, such as books, articles, websites, and social media. They can capture natural language's semantic and syntactic patterns and relationships [19,29]. While LLMs have driven advances in text generation, generative AI broadly refers to various techniques for synthesizing novel data or content across modalities. Beyond natural language, generative models can produce images, 3D models, audio, video, and more.

## 2.1. Foundational algorithms

The foundational algorithms of generative AI are the core methods and techniques that enable the generation of data or content. These algorithms can be categorized into different types based on underlying principles and assumptions.

**Figure 2** provides a schematic representation of these algorithms and their functionalities in generating/encoding, decoding/reconstructing/translating texts, images, and audio.

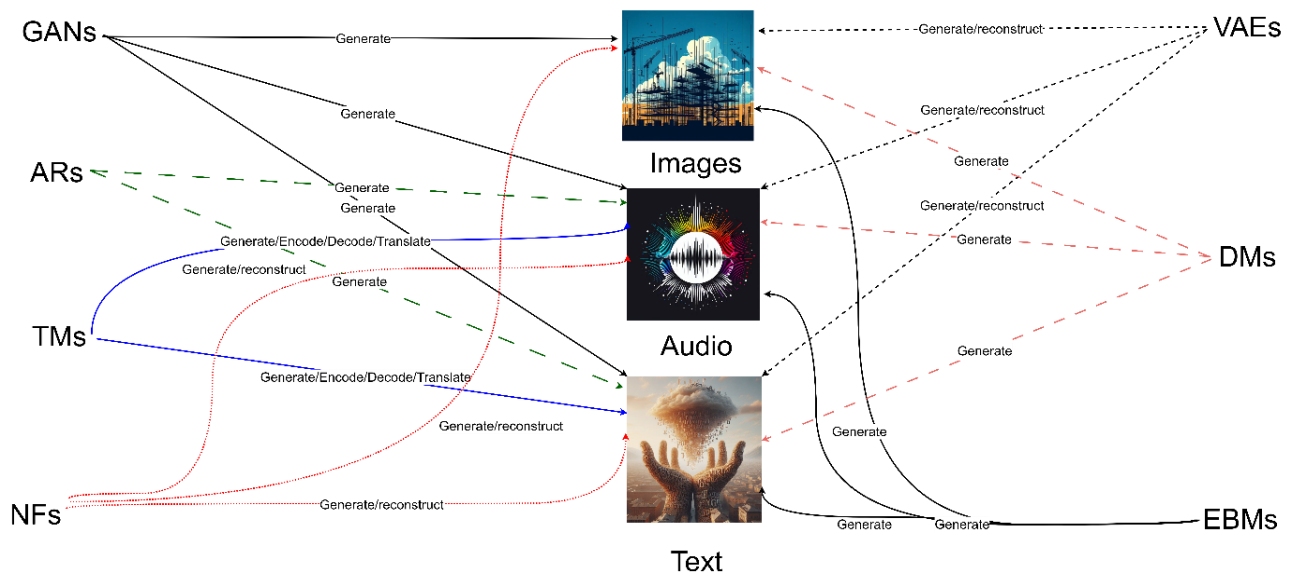

**Figure 2:** Schematic summarizing the foundational algorithms for generating/encoding, decoding/reconstructing/translating images, text, and audio. GANs (generative adversarial networks), ARs (autoregressive models), VAEs (variational autoencoders), DMs (diffusion

models), TMs (transformer models), NFs (normalizing flows), and EBMs (energy-based models).

### 2.1.1. Generative adversarial networks (GANs)

GANs are generative AI algorithms that use a game-theoretic approach to generate data or content by competing between a generator and a discriminator. The generator tries to produce realistic and diverse data or content, while the discriminator tries to distinguish between the real and the fake data or content [34]. GANs can generate data or content, such as images, videos, and audio, based on some input or noise. Still, they can suffer from mode collapse, where the generator produces limited variations of data or content. **Table 1** shows common GAN-based models and their functions.

**Table 1: GAN-based models and their functions**

| Generative adversarial networks | Function | Ref. |
|---|---|---|
| StyleGAN | A model that can generate high-quality and diverse images of human faces | [35] |
| CycleGAN | A model that can translate images from one domain to another without paired data | [36] |
| BigGAN | A model that can generate large-scale and high-fidelity images of various classes | [37] |

### 2.1.2. Autoregressive models (AMs)

AMs are a generative AI algorithm that uses a sequential approach to generate data or content by predicting the next element based on the previous elements. The model learns the conditional probability distribution of the data or content and its samples to generate new data or content. Autoregressive models can generate sequential and structured data or content, such as text, audio, and code, based on some input or context. A limitation is exposure bias - during training, AMs only see ground truth sequences, not their predictions. So errors can accumulate at inference [38]. **Table 2** presents examples of AM-based models and their key capabilities.

**Table 2: Autoregressive models and their functions**

| Autoregressive models | Function | Ref |
|---|---|---|
| PixelRNN | A model that can generate realistic images pixel-by-pixel | [39] |

| WaveNet | A model that can generate realistic speech and music waveforms | [40] |
| GPT-3 | A model that can generate natural language text for various tasks | [41] |

### 2.1.3. Variational autoencoders (VAEs)

VAEs rely on a probabilistic approach for generating novel data or content. These models map inputs into a latent space and then sample from a prior distribution. VAEs can produce varied outputs that resemble the original data by learning to minimize the reconstruction error and divergence between prior and posterior distributions. Images, videos, audio, and other modalities can be generated by sampling from the latent space based on random noise or conditional inputs [42]. However, VAEs face the challenge of posterior collapse, where the latent space fails to capture meaningful variational information to condition the generations. Some VAE-based models are presented in **Table 3**.

**Table 3: VAE-based models and their functions**

| Variational autoencoders | Function | Ref |
| --- | --- | --- |
| VAE | A model that can learn a latent representation of data and generate new data | [42] |
| CVAE | A model that can learn a conditional latent representation of data and generate new data | [43] |
| VQ-VAE | A model that can learn a discrete latent representation of data and generate new data | [44] |

### 2.1.4. Transformer models (TMs)

Transformers represent a paradigm shift in generative modeling underpinned by attention mechanisms. Rather than sequence alignment, self-attention layers allow modeling long-range dependencies in data by learning correlations between input and output vectors. This gives transformers a global receptive field for generation compared to RNNs' localized windows [29,45]. Beyond sequences like text and audio, transformers can generate graphs, 3D points, and other structural data using self-attention. However, challenges remain. Without explicit alignment, representing order and continuity is difficult. Catastrophic forgetting of rare sequences also occurs as new training data overrides previously learned patterns. Restricting self-attention to local neighborhoods may improve stability [46]. **Table 4** presents common transformer-based models and their functions.

**Table 4: Transformer-based models and their functions**

| Transformer models | Function | Ref. |
|---|---|---|
| GPT-4 | A model that can perform sequence-to-sequence tasks using attention mechanisms | [47] |
| BERT | A model that can learn bidirectional representations of natural language for various tasks | [48] |
| ViT | A model that can learn visual representations of images using transformers | [49] |

### 2.1.5. Diffusion models (DMs)

Diffusion models generate data through a two-step denoising process. First, the algorithm adds isotropic Gaussian noise to the real data over repeated diffusion steps. This gradually destroys structure while maintaining dimensionality. Next, the process is reversed by a neural network that removes noise step-by-step until pristine samples emerge. Unlike autoregressive methods, diffusion models generate entire data instances simultaneously rather than sequentially [50]. This provides inherent parallelism. However, determining optimal diffusion and denoising schedules remains challenging. Models must also undo all noise perfectly or risk compounding errors. Slow convergence arises from the numerous forward and reverse passes required. Recent innovations like denoising score matching have accelerated diffusion model training [51]. Diffusion-based models are presented in **Table 5**.

**Table 5**: Diffusion-based models

| Diffusion models | Function | Ref. |
|---|---|---|
| DDPM | A model that can generate realistic images by reversing a diffusion process | [52] |
| NCSN | A model that can learn the score function of data distribution using noise-conditional score networks | [53] |
| DALL-E 2 | A model that can generate images from text captions using the diffusion technique | [54] |

## 2.1.6. Normalizing flows (NFs)

NFs offer an invertible approach to density estimation for generative modeling. Applying a series of bijective mappings can transform data into a latent space where the prior distribution is known [55]. This allows exact likelihood calculation. The change-of-variables formula relates probabilities between spaces through the Jacobian. Normalizing flows learn mappings that maximize the likelihood of the training data under the latent space prior. Sampling new points is achieved by testing the latent prior and reversing the flow. Normalizing flows shine where data lies near a lower-dimensional manifold, not the entire ambient space. However, computational complexity rises linearly with depth due to repeated Jacobian determinants. Trade-offs exist between modeling power and efficiency. Recent work has enhanced flows with concepts like sparsity, conditional inputs, and new invertible layers. Normalizing flows remain promising for manifold learning-based generation [55]. **Table 6** outlines examples of NF-based models and their key capabilities.

**Table 6:** NF-based models and their functions

| Normalizing flows | Function | Ref. |
|---|---|---|
| NICE | A model that can learn a bijective mapping between data and noise using additive coupling layers | [56] |
| RealNVP | A model that can learn a bijective mapping between data and noise using affine coupling layers | [55] |
| Glow | A model that can learn a bijective mapping between data and noise using invertible 1x1 convolutions and actnorm layers | [57] |

## 2.1.7. Energy-based models (EBMs)

EBMs take a thermodynamics approach to generative modeling by formulating data density as Boltzmann distribution. Lower energies indicate higher probability density. The architecture consists of an energy function assigning scalar values to each data configuration and a sampling procedure to generate new data. Typically, the energy function is represented by a deep neural network like a convolutional autoencoder. The network is trained to assign low energies to observed training examples and higher energies elsewhere. Sampling generates new data by initializing random noise and descending via gradient descent until reaching local energy minima. A significant challenge is mode collapse, where sampling needs to cover the full diversity of densities. Insufficient capacity in the energy network also hinders quality. Modern advances integrate EBMs with MCMC sampling and discriminator networks to improve coverage and sample quality [58]. The physics-inspired energy framework provides a unique generative modeling perspective differing from prevailing probabilistic approaches. **Table 7** shows common energy-based models with their capabilities.

**Table 7**: Energy-based models and their functions

| Energy-based models | Function | Ref. |
|---|---|---|
| RBM | A model that can learn a joint distribution of data and hidden variables using a bipartite graph | [59] |
| Hopfield network | A model that can store and retrieve patterns using a recurrent network | [60] |
| EBGAN | A model that can generate realistic images using an energy-based discriminator | [58] |

## 2.2. Large generative models

Large generative models (LGMs) refer to massive neural networks trained on huge datasets that can generate text, images, audio, video, or code. By exposing the model to large corpora of varied content during pretraining, LGMs learn general representations that empower them to generate original outputs within specific modalities. These models are powered by the foundational algorithms discussed in the previous section. For example, models like GPT-4 and Codex can produce coherent text and functional code based on prompts, respectively [27,29]. DALL-E 3 and Imagen can generate photorealistic images from text descriptions [51,61]. Models trained on audio can synthesize natural human speech or music. LGMs can also combine modalities to enable multimodal generation, such as illustrating input stories with suitable images or accompanying lyrics with fitting music compositions. The defining feature of LGMs is their massive scale in model size, computational requirements, and training data volume, which enables versatile and creative generation spanning from text to images to video to code and in an integrated multimodal fashion. This makes them a multipurpose generative toolbox powered by pretraining on diverse big data. Table 9 lists LGMs released in recent years, their developer, training parameters, release year, and accessibility.

**Table 8:** LGMs released in recent years.

| LGM | Developer | Training parameter (Billion) | Release year | Access | Reference |
|---|---|---|---|---|---|
| GPT-4 | OpenAI | 1000 | 2023 | API | [62] |
| Gemini Pro | Google DeepMind | 17 | 2023 | Open source | [63] |
| Llama 2 | Meta | 2000 | 2023 | Open source | [64] |
| PaLM | Google | 540 | 2022 | Open source | [65] |
| Claude | Anthropic | 12 | 2023 | Open source | [66] |
| DALLE-3 | OpenAI | - | 2023 | API | [51] |
| SDXL | Stability AI | 2.6 | 2023 | Open source | [67] |
| DALLE-2 | OpenAI | 3.5 | 2022 | API | [68] |
| Dreamfusion | Google | - | 2021 | Open source | [69] |
| Flamingo | Google DeepMind | 3.2 | 2022 | API | [70] |
| Phenaki | Google | 1.8 | 2022 | API | [71] |
| Codex | OpenAI | 100 | 2021 | API | [72] |
| Galactica | Meta | 120 | 2022 | API | [73] |
| AudioLM | Google | 0.6 | 2022 | API | [74] |
| AlphaTensor | DeepMind | 0.5 | 2021 | Open source | [75] |
| DALL-E | OpenAI | 12 | 2021 | API | [76] |
| CLIP | OpenAI | 63 | 2022 | Open source | [77] |
| BART | Facebook | 0.4 | 2019 | Open source | [78] |
| T5 | Google | 11 | 2019 | Open source | [79] |
| BERT | Google | 0.34 | 2018 | Open source | [48] |
| GPT-3.5 | OpenAI | 175 | 2022 | API | [80] |
| GPT-2 | OpenAI | 1.5 | 2019 | Open source | [81] |
| XLNet | Google | 0.34 | 2019 | Open source | [74] |

## 3. Methodology

A four-phase approach is adopted to achieve the objectives of this study. Figure 1 visualizes these phases, including systematic literature review and retrieval, expert discussion and review, a framework for developing custom LGM in the construction industry, and a case study.

I. **Phase 1 – Systematic literature retrieval and review**: The first step in this phase involves selecting appropriate databases for the literature search. Scopus, Web of Science, and ScienceDirect were chosen due to their broad coverage and rigorous indexing of peer-reviewed publications [82,83]. Keyword identification was conducted iteratively to capture relevant studies at the intersection of generative AI and the construction industry. The final search string consisted of ["Construction industry" OR "architecture engineering and construction industry" OR "AEC industry" OR "AECO industry"] AND ["Generative AI" OR "GenAI" OR "GENAI" OR "Bard" OR "Gemini" OR "GPT" OR "GPT-1" OR "GPT-2" OR "GPT-3" OR "InstructGPT" OR "ChatGPT" OR "Transformer" OR "GPT-4" OR "Llama" OR "LamDA"]. This search returned 79 initial results. We narrowed the search results to 10 potentially relevant studies based on the title and abstract screening. An in-depth review found that only four (4) were original research articles, with two review articles, and the full text of the rest was unavailable or written in languages other than English. Snowball searching expanded the final pool to six (6) peer-reviewed papers at the intersection of generative AI and construction.

II. **Phase 2 – Expert discussion and review**: The limited literature identified in Phase 1 highlighted the need to supplement with expert perspectives, given the nascent state of generative AI adoption in construction. To elicit diverse insights, 15 experts with backgrounds spanning AI research and construction industry practice were identified. Invitations were sent to participate in the study, with 11 experts accepting for a 73% response rate. This panel encompassed university professors in AI and construction engineering, technology directors from major construction firms, and founders of AI startups targeting the architecture, engineering, and construction (AEC) industry. A modified Delphi survey was conducted with the panel to identify opportunities and challenges of applying generative AI in the construction industry. Thematic analysis was then used to extract common themes from the qualitative responses. This involved codifying the experts' opinions and aggregating them into categories through an iterative process. The goal was to determine areas of consensus as well as unique perspectives.

III. **Phase 3 – Framework for developing custom LGM in the construction industry**: This phase involved synthesizing the literature and expert findings into a methodology construction firms can follow to build custom generative AI solutions using their proprietary data. The framework encompasses construction data collection, dataset curation, training the custom LGM, evaluation, and deployment steps.

IV. **Phase 4 – Case study**: A case study was conducted using generative AI for querying contract documents to demonstrate practical application. The first step involved the selection of the base LLM architecture. OpenAI's GPT-4 model was chosen as the base model due to its state-of-the-art natural language generation capabilities. A retrieval-augmented generation (RAG) system was implemented to improve the base LLM further. This mitigated hallucinated text by grounding outputs in relevant dataset examples. LangChain Library was employed for the development [84]. The performance of the customized LLM was evaluated, and a graphical user interface was developed using Streamlit [85]. This interactive web application enabled testing of the customized generative AI model through prompts.

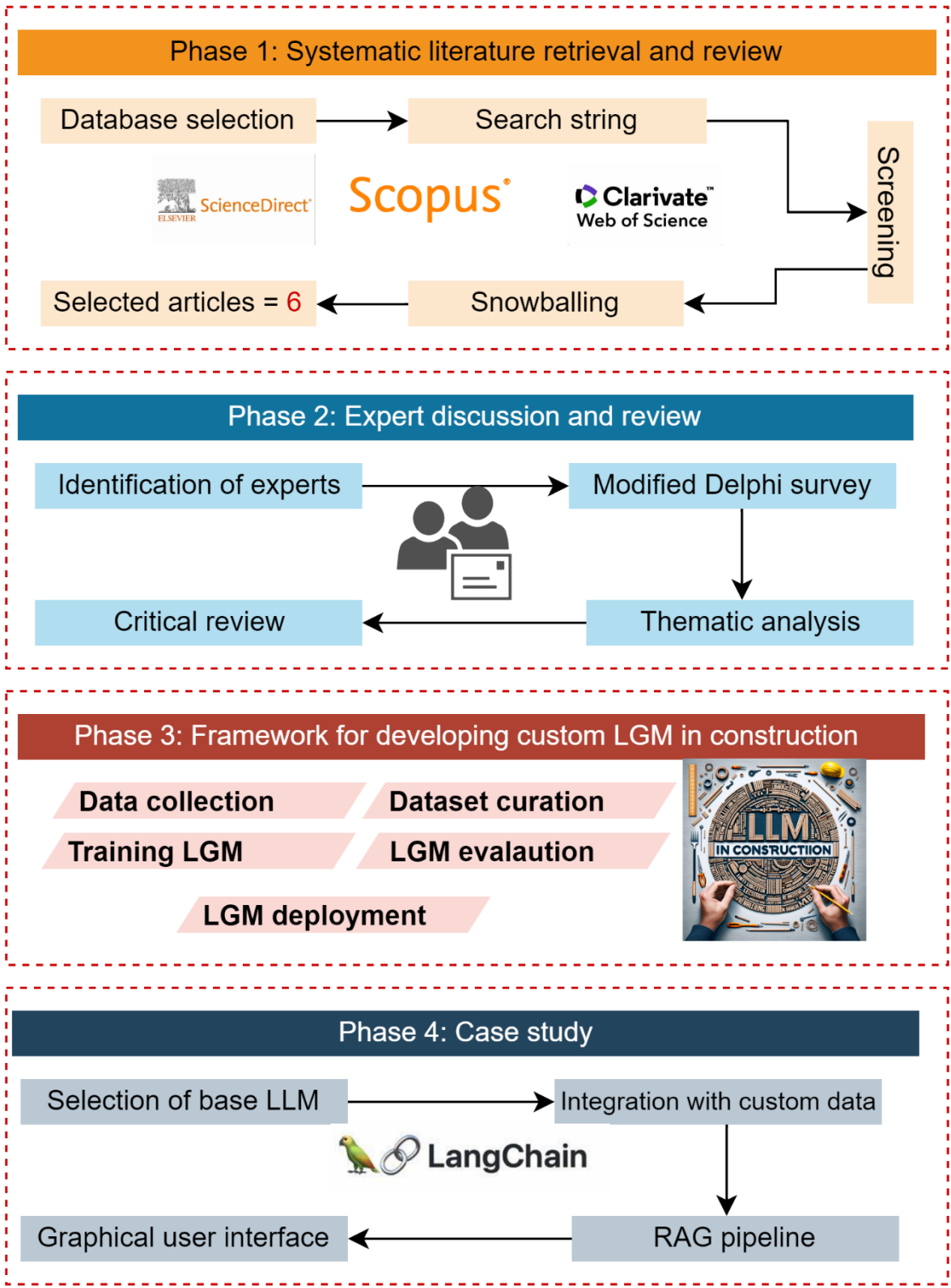

**Figure 3:** Research framework

# 4. Results

## 4.1. Current applications

This section presents the application of generative AI in the form of LGMs in the construction industry. Based on the systematic review, only six peer-reviewed articles exploring the uses of LLMs in construction were identified. No articles relating to other LGMs, such as large image and video models, were found. The six articles are summarized in **Table 10**, including their objective, methods, and contributions. The reviewed studies demonstrate emerging applications of LLMs, such as GPTs and BERT-based models for construction tasks, including virtual assistance, sequence planning, schedule generation, hazard recognition, risk assessment, and project planning [86–88]. The contributions highlight the potential for LLMs to enhance productivity, accuracy, and automation in areas like information retrieval, education/training, and documentation review. However, the limited number of studies indicates that the adoption of modern generative AI in construction is still in the very early stages. Significant research is needed to develop customized LLMs for the industry and validate their capabilities on real-world problems.

Table 9: Summary of current applications of LLM in the construction industry

| Reference | Objective | Methods | Contributions |
|---|---|---|---|
| [89] | Development of a dynamic prompt-based virtual assistant framework for BIM information search | The framework integrates BIM and GPT technologies for an NL-based interface.<br><br>Dynamic prompt-based process interprets NL queries, retrieves information, and delivers responses. | The framework's application improves information search speed, accuracy, and user experience. |
| [90] | Development of RobotGPT for automated sequence planning in robotic assembly for construction tasks. | RoboGPT is a system that uses ChatGPT for automated sequence planning in robot-based construction assembly.<br><br>The experimental evaluation included two case studies and 80 trials involving real construction tasks. | RoboGPT-driven robots can handle complex construction operations and adapt to changes on the fly. |
| [91] | Generation of a construction schedule for a project | ChatGPT is employed to generate a construction schedule for a simple project.<br><br>A survey was conducted to evaluate output quality and participants' experience.<br><br>Parameters used to evaluate results include accuracy, efficiency, clarity, coherence, reliability, relevance, consistency, scalability, and adaptability. | The use of LLM to enhance construction schedules workflow. |
| [92] | The use of ChatGPT for improving hazard recognition on construction site | The investigation involved 42 students in a construction program.<br><br>Pre- and post-intervention hazard recognition abilities were measured. | The potential of employing ChatGPT for safety education and training. |
| [93] | Automated classification of contractual risk clauses | The BERT method is used for clause classification in construction specifications.<br><br>Seven risk categories were identified: payment, temporal, procedure, safety, role and responsibility, definition, and reference. | The model improves the construction specification review process and risk management. |
| [94] | Automatic matching of look-ahead planning tasks to master scheduled activities | Both location-based and distance-based matching followed were employed.<br><br>GPT-2 was used for final matching. | Auto-alignment of long-term and short-term plans in construction projects |

## 4.2. Opportunities

The discussions conducted with experts revealed numerous potential opportunities to deploy generative AI across construction, categorized by input-output capabilities. Sections 4.2.1 through 4.2.9 extensively examine applications that generate text, images, and video. While generative models can also synthesize audio, experts advised that video generation can serve dual visual and auditory content purposes as construction relies heavily on visual data like drawings, photos, animations, and written and verbal communications, generative modes spanning text, images, and video were seen as most directly relevant. Despite significant progress, there are still limitations, particularly when generating complex images and videos. **Table 11** summarizes leading generative models for different data types. LLMs like GPT-4 and Gemini Pro demonstrate proficiency in text synthesis [45,95]. DALL-E 3 produces images from text captions [51]. Video generation models like CogVideo, Lumiere, and Stable Diffusion show promise but are still being refined [96,97]. Although there are shortcomings, the pace of progress makes generative AI a promising technology for transforming the construction industry. If trained on sufficient domain data, Text generation achieves high coherence and accuracy. Photorealistic image synthesis provides value in design and documentation use cases. Video capabilities lag but rapidly improve through advances like higher resolution GANs [97,98].

**Table 10:** Generative AI models for various input-output types

| Input-output type | Model | Developer | Reference |
|---|---|---|---|
| Text to text | GPT-3, GPT-4, Gemini Pro | OpenAI, Google's DeepMind | [45,95] |
| Text to image | DALL-E 3 | OpenAI | [51] |
| Text to video | CogVideo, Lumiere | Nightmareai, Google Research | [96,97] |
| Image to text | GPT-4, Gemini Pro | OpenAI, Google's DeepMind | [45,95] |
| Image to image | Pix2Pix | Berkeley AI Research | [99] |
| Image to video | Stable Video Diffusion | Stability AI | [50] |
| Video to text | VideoCoCa | Google Research | [100] |
| Video to image | - | - | |
| Video to video | Lumiere, Gen-2 | Google Research, Runway | [97,98] |

4.2.1. Text to text

Generative AI revolutionizes the construction industry by converting textual data into advanced textual outputs, assisting in many tasks in the construction project phases. **Table 12** provides a detailed overview of the potential applications of text generation in the construction industry, categorized using the different project phases. In pre-construction, it can help generate feasibility study summaries, ensure regulatory compliance, and automate proposal/bid drafting [101]. Drafting daily progress reports, specifications, task instructions, and other documents can be automated during construction (see **Figure 4**). Post-construction opportunities include creating inspection reports, punch lists, operation and maintenance manuals, reviewing warranty/compliance letters, and translating documents. Other cross-cutting text applications are information retrieval through natural language queries and translation into multiple languages [16]. With proper training in technical corpora, they can translate industry insights directly into clear, accurate documents without tedious hands-on work. Realizing this potential requires careful, prompt engineering and alignment with construction linguistic patterns and technical jargon.

**Table 11:** Potential generative AI opportunities in the construction industry for text-text model type

| Potential opportunity | Description | Project phase |
|---|---|---|
| Generation of the feasibility report summary | Summarize extensive feasibility reports and extract key insights and recommendations for informed decision-making during the project initiation. | Pre-construction |
| Documentation of regulatory compliance | Leverage generative AI to assist in creating documents that ensure compliance with regulatory requirements, a crucial task in the pre-construction planning phase. | Pre-construction |
| Preparation of proposal/bid | Apply generative AI to assist in the preparation of proposals and bids by automatically generating well-structured and persuasive text content. | Pre-construction |
| Generation of daily progress report | Create a daily progress report template summarizing on-site activities and achievements during construction. | Construction |
| Refinement of construction specifications | Utilize generative AI to refine and enhance construction specifications, ensuring clarity and accuracy in the documentation of materials, methods, and standards. | Construction |
| Task Assignment and Communication | Facilitate task assignment and communication by automatically generating clear and detailed instructions for construction teams through generative AI. | Construction |
| Summarization of as-built documents | Summarize the extensive as-built documentation, providing a condensed overview of the final constructed project for post-construction analysis. | Post-construction |
| Generation of facility maintenance manual | Automate the generation of comprehensive facility maintenance manuals based on the final as-built documentation. | Post-construction |
| Review of warranty and compliance document | Utilize generative AI to review warranty and compliance documents, summarizing critical information and ensuring adherence to post-construction requirements. | Post-construction |
| Language translation and localization | Translate text content between different languages, aiding in global collaboration and communication. | All |
| Information retrieval and knowledge discovery | Enhance contextual search capabilities by using generative AI to understand and respond to natural language queries, improving the accuracy of information retrieval. | All |

**You**
Create a comprehensive template for construction daily progress report

**ChatGPT**
Certainly! Below is a comprehensive template for a construction daily progress report:

# Construction Daily Progress Report

**Project Name:** [Project Name]
**Date:** [Date]
**Weather Conditions:** [Weather Conditions]

## 1. Project Overview:

| | |
|---|---|
| **Project Location:** [Location] | **Project Duration:** [Start Date] to [End Date] |
| **Project Manager:** [Project Manager's Name] | **Contractor:** [Contractor's Name] |
| **Client:** [Client's Name] | **Project Phase:** [Phase of the Project] |

## 4. Equipment and Material Updates:

### A. Equipment:

| Equipment Used Today | Equipment Maintenance | Equipment Availability |
|---|---|---|
| [List of equipment used] | [Any maintenance or repair work carried out] | [Note if any equipment is unavailable or requires servicing] |

### B. Materials:

| Materials Received Today | Materials Used | Material Shortages |
|---|---|---|
| [List of materials received] | [Specify materials used and quantities] | [Any shortages or delays in material delivery] |

## 5. Health and Safety:

### A. Incidents:

- **Accidents/Incidents Today:** [Report any accidents or incidents]
- **Safety Violations:** [Note any safety violations observed]

## 2. Summary of Activities:

### A. Work Completed Today:

- [Brief description of tasks completed]
- [Details of completed work, including milestones achieved]
- [Any challenges encountered and how they were addressed]

### B. Work Planned for Tomorrow:

- [Tasks planned for the next working day]
- [Equipment/material deliveries scheduled]
- [Personnel assignments]

## 3. Progress Photos:

### A. Completed Work:

- [Attach photos of completed work]
- [Include captions describing each photo]

### B. Work in Progress:

- [Attach photos showing ongoing work]
- [Provide explanations or captions if necessary]

### B. Safety Measures:

- **Safety Inspections:** [Details of safety inspections conducted]
- **Safety Briefings:** [Any safety briefings held]

## 6. Remarks and Additional Notes:

- [Any additional remarks or comments]
- [Notes on any outstanding issues or concerns]

## 7. Next Steps:

- [Outline next steps and upcoming tasks]
- [Any changes or adjustments to the project schedule]

## 8. Signatures:

**Project Manager:** ___________________________
**Date:** ___________________________

**Figure 4:** Daily progress report template generated using GPT-4 via ChatGPT interface

4.2.2. Text to image

AI's text-to-image conversion provides innovative possibilities in the construction field, including the ability to visualize pre-construction architectural ideas, assist in making real-time construction choices, and enhance marketing materials once construction is completed. The potential opportunities for generating images via text prompting are shown in **Table 13**. Generating images from text has broad applicability in construction projects. Pre-construction applications include creating visualizations from site descriptions for selection and planning [102]. Text prompts can also render architectural concepts and project models (**Figure 5**). During construction, progress visualization, equipment layouts, and safety illustrations can be automated from textual inputs. **Figure 6** shows a visualization of construction progress through different stages of execution. Post-construction use cases involve as-built visualization, usage guidelines, and renovation proposals. With appropriate training, models like DALL-E can translate construction domain language into detailed visuals through well-prompted texts. This technology allows people without expertise to readily obtain visual depictions by articulating what they wish to see in plain language. Automating this linkage between vision and language can make project information more accessible while freeing worker time [103].

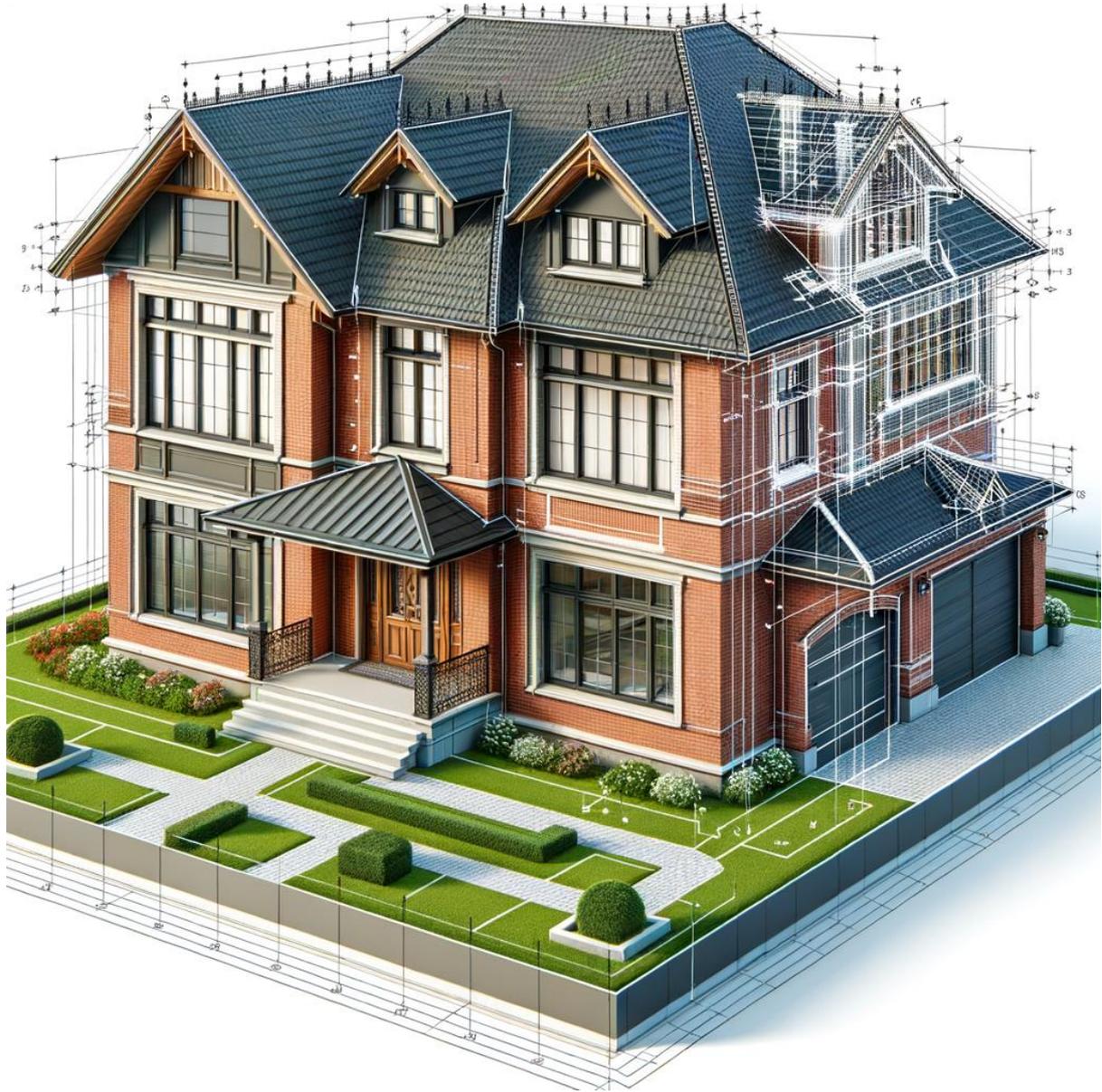

**Figure 5:** 3D BIM Model Generated Using DALL-E 3

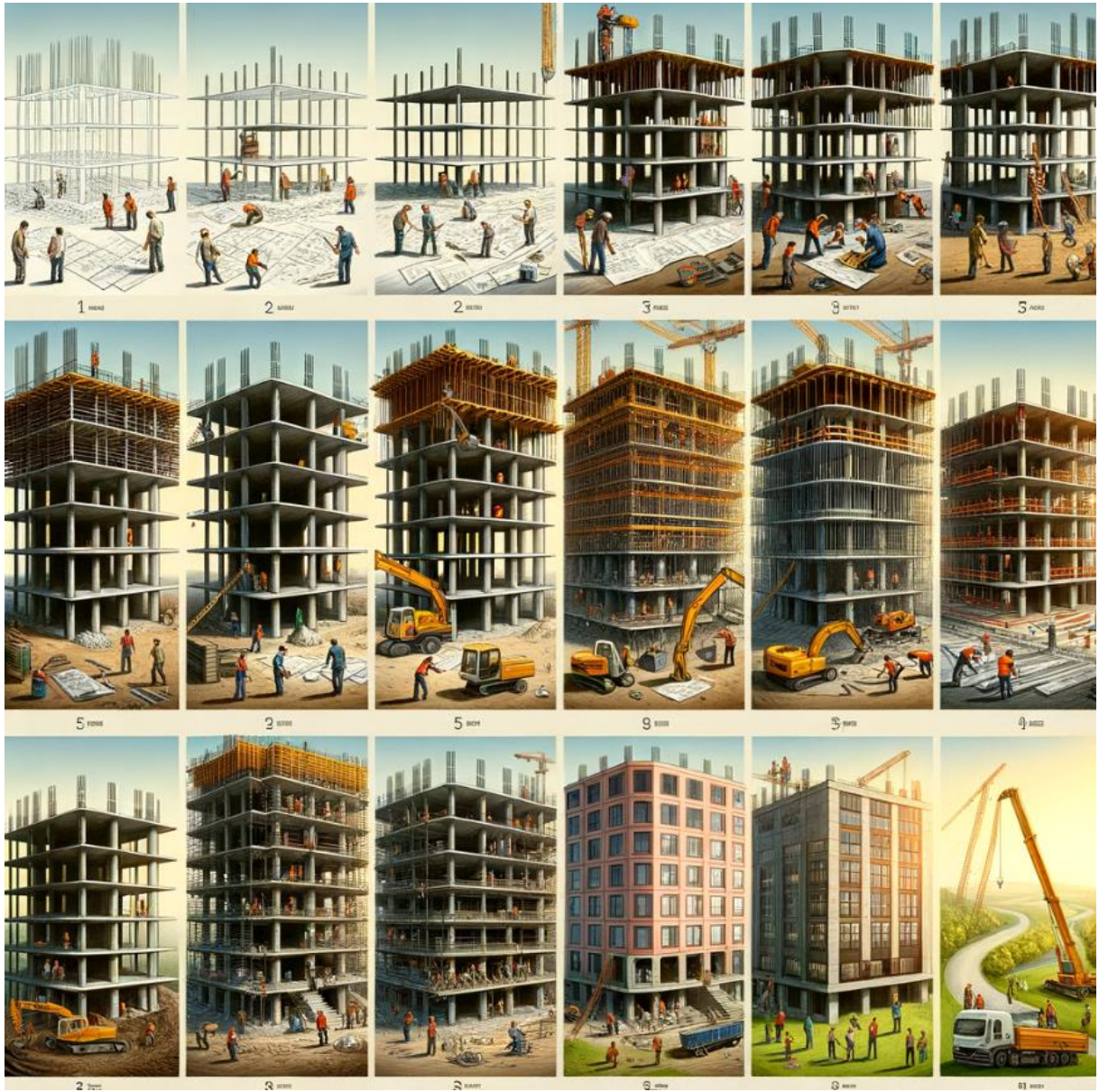

**Figure 6:** Construction progress visualization generated by DALL-E-

Table 12: Potential generative AI opportunities in the construction industry for text-image model type

| Potential opportunity | Description | Project phase |
|---|---|---|
| Site visualization and selection | Create visual representations of potential construction sites based on text descriptions, aiding the decision-making process during site selection. | Pre-construction |
| Architectural concept rendering | Transform textual architectural concepts into visual renderings, providing stakeholders with a clear preview of the proposed designs. | Pre-construction |
| Interactive project models | Utilize generative AI to convert project descriptions into interactive 3D models, allowing stakeholders to explore and engage with the project before construction begins. | Pre-construction |
| Construction progress visualization | Implement generative AI to generate visual representations of construction progress based on textual updates, providing stakeholders with a visual timeline of the project. | Construction |
| Material and equipment layouts | Through generative AI, create visual layouts of materials and equipment based on textual descriptions, optimizing their placement on the construction site. | Construction |
| Safety procedure illustrations | Apply generative AI to convert text-based safety procedures into visual illustrations, enhancing comprehension and adherence to safety protocols on the construction site. | Construction |
| As-Built visualization | Transform the as-built documentation into visual representations, aiding in the visualization and analysis of the final construction. | Post-construction |
| Facility usage guidelines | Create visual guidelines for facility usage based on textual documentation, ensuring clear communication of post-construction guidelines. | Post-construction |
| Renovation proposal visualizations | Generate visual representations of proposed renovations, aiding decision-making during the post-construction phase. | Post-construction |
| Project timeline infographics | Convert textual project timelines into visual infographics, providing an easily understandable overview for all project phases. | All |
| Project dashboard visuals | Generate visual representations for project dashboards based on textual data, offering stakeholders an intuitive and informative overview of project metrics. | All |

### 4.2.3. Text to video

The utilization of generative AI to transform textual information into dynamic video content offers numerous benefits. **Table 14** summarizes key opportunities for text-to-video generation in construction based on the expert discussion. During the pre-construction phase, introductory site exploration videos and animated project concept videos could be synthesized from text to aid scope planning and stakeholder intelligence. During construction, step-by-step equipment operation tutorials and safety training animations could be generated from manuals and textual hazard narrations, respectively [104]. Progress update videos compiled from schedules and logs would help keep stakeholders informed with matching visual updates. Post-construction use cases include creating instructional facility usage videos from the documentation. With appropriate training data, text-to-video models can translate construction domain language into vivid animations and live footage [96,97]. Rather than relying solely on static diagrams and dense text, bringing instructions and processes to life through AI-generated videos makes project information more engaging. Dynamic video tutorials personalized via text to each situation may enhance comprehension and learning for safety training and equipment operation. Automating the linkage between textual descriptions and video footage also frees workers time spent manually storyboarding and editing visualizations. As text-to-video generation techniques continue advancing in resolution and realism, the applications across the construction project lifecycle will expand.

**Table 13:** Potential generative AI opportunities in the construction industry for text-video model type

| Potential opportunity | Description | Project phase |
|---|---|---|
| Site introduction videos | Create introductory videos for potential construction sites, providing stakeholders with visual overviews based on textual descriptions. | Pre-construction |
| Project concept animation | Transform textual project concepts into animated videos, offering stakeholders a dynamic visualization of the proposed construction. | Pre-construction |
| Equipment operation guides | Generative AI can automatically create step-by-step video tutorials demonstrating equipment use from the text and diagrams in instruction manuals. | Construction |
| Safety procedure animations | Safety managers could compose comprehensive narrations of hazards and precautions. Generative AI can synthesize engaging video footage matching the narration to create safety training materials. | Construction |
| Progress update videos | Generate progress videos automatically using generative AI using progress reports, schedules, logs, and notes. | Construction |
| Facility usage instruction videos | Generate instructional videos based on textual documentation for facility usage, ensuring clear communication of post-construction guidelines. | Post Construction |
| Building update videos | Produce AI-generated videos summarizing facility modifications, upgrades, and status changes over time from text-based building logs for stakeholders. | Post Construction |
| Project journey montage | Implement generative AI to compile a video montage showcasing the entire project journey, combining textual descriptions and visual elements for a comprehensive overview. | All |

4.2.4. Image to text

Applying AI in converting images into text significantly improves construction procedures at all stages. Converting images into text descriptions also has valuable applications throughout construction project phases. **Table 15** summarizes potential use cases validated by the expert discussion. Pre-construction opportunities include extracting measurements, boundaries, and other information from land surveys and blueprints. During construction, daily site photos could be analyzed to generate progress reports. Images of materials and equipment could develop real-time quality and inventory assessments via generative AI [105]. Figure 7 displays an image description generated by Gemini Pro as part of a daily visual report. Upon close inspection, the model accurately captures fine-grained details in the image, including identifying the specific brand and model of the construction equipment. This demonstrates Gemini Pro's capability to produce descriptive text summarizing critical visual information [95]. Post-construction applications involve extracting as-built details from archival photos and making warranty documentation from damaged images. Cross-cutting use cases include automating visual inspection reports across phases. With proper training, image captioning techniques can translate construction graphics and photos into structured textual information. This eliminates tedious manual efforts to log and convey visual observations. Models such as GPT-4 can analyze everyday images and accurately describe prominent objects, actions, and scenery [45].

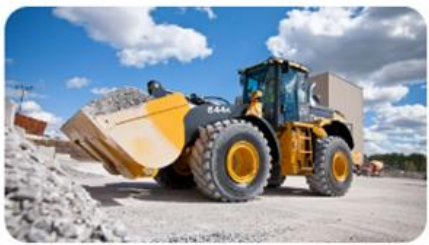

Figure 7: Description of an image taken as part of a daily visual report generated by Gemini Pro

**Table 14:** Potential generative AI opportunities in the construction industry for image-text model type

| Potential opportunity | Description | Project phase |
| --- | --- | --- |
| Land survey data extraction | Analyze land survey images and extract textual data, such as measurements, topographical details, and boundary information. | Pre-construction |
| 3D model specification | Analyze 3D architectural models and automatically generate detailed textual specifications of materials, components, dimensions, etc. | Pre-construction |
| Blueprint digitization | Automatically convert scanned paper blueprints and hand-drawn sketches into structured digital representations. | Pre-construction |
| Daily progress image analysis | Analyze daily progress images from construction sites and generate textual reports summarizing the progress, challenges, and achievements. | Construction |
| Material quality assessment | Examine images of construction materials and generate textual assessments regarding quality, potential issues, and compliance. | Construction |
| Inventory management | Use AI techniques to automatically catalog on-site equipment, materials, tools, etc., from images and videos into searchable inventory databases. | Construction |
| As-built documentation text extraction | Analyze images of as-built documentation and extract textual information, facilitating the creation of detailed post-construction reports. | Post-construction |
| Warranty claim documentation | Explore images of construction components and generate textual documentation for warranty claims, specifying issues and relevant details. | Post-construction |
| Visual inspection reports | Examine images from visual inspections and generate textual reports, providing detailed information on observed conditions and recommendations. | All |

4.2.5. Image to image

The image-to-image capabilities of generative AI are crucial in construction as they facilitate design adjustments, on-site issue solutions, and the visualization of future upgrades. **Table 16** details potential applications of image-image models in the construction industry. Pre-construction use cases include adapting architectural visualizations into different desired art styles and refining scanned maps into clear site plans. During construction, input architectural plans and sketches could be auto-modified to match ongoing changes on-site [28]. Material texture libraries could help generate realistic composite renderings from sample images. Post-construction applications involve visualizing the restored building state from damage assessment images and landscape enhancements. Improving image quality and resolution are potential applications across all phases of construction projects. Techniques such as pix2pix GANs demonstrate capabilities to transform input images while preserving essential content structure [99]. By learning alignments between construction image domains during training, models can translate inputs into desired stylistic, structural, or conceptual outputs. This allows the adaptation of visual data into appropriate formats for downstream usage, reducing repetitive manual editing. For instance, rough sketches produced during early design phases can be refined into polished architectural visualizations or engineering schematics. Images captured on-site can be adapted to match design intent, even when physical conditions vary. Continued advances in high-resolution GANs will further expand the potential for image-to-image synthesis to enhance visual media throughout construction projects.

**Table 15:** Potential generative AI opportunities in the construction industry for image-image model type

| Potential opportunity | Description | Project phase |
| --- | --- | --- |
| Architectural image translation | Use generative techniques to adapt architectural visualizations and renderings done in one style to different target art styles. | Pre-construction |
| Site planning refinement | Refine scanned maps and satellite imagery to generate clear site/lot diagrams and top-down site plans for planning. | Pre-construction |
| Concept generation | Produce variations of initial architectural sketches and concept art to explore broader design possibilities. | Pre-construction |
| Updating architectural drawing | Modify architectural drawings and plans by incorporating changes made on the construction site to keep documentation up-to-date. | Construction |
| Material texture matching | Apply generative AI to match the textures of construction materials with reference images, ensuring consistency and quality in the visual appearance of the constructed elements. | Construction |
| Damage assessment | Process images of damaged building areas and generate visualizations showing restored states. | Post-construction |
| Landscape transformation visualization | Visualize the transformation of landscapes based on input images, supporting post-construction projects such as garden enhancements or environmental modifications. | Post-construction |
| Aesthetic enhancement | Improve resolution, lighting, orientation, and low-quality construction images across phases. | All |

4.2.6. Image to video

Converting static images into dynamic videos opens up impactful possibilities across the construction project lifecycle. **Table 17** summarizes potential applications in this area. During pre-construction, static architectural concept images could be converted into engaging animated walkthroughs and fly-throughs to showcase designs. Aerial site photos could also produce simulated planning and development timelapses [106]. Safety incidents could be recreated on active construction sites based on analysis of images of unsafe conditions to improve hazard awareness through vivid video representations. Timelapse build videos compiled from daily construction photos help visually track project progression. Post-construction use cases include generating promotional experience videos from facility images and collecting recap documentary videos from archival visuals. State-of-the-art generative video models demonstrate increasing capabilities to animate photo-realistic footage from sparse image inputs [50]. AI systems can extend single images into complete video sequences with convincing continuity and realism by learning to extrapolate motion and physical interactions [107]. Construction visuals contain extensive intrinsic structures that video generation models can leverage to produce meaningful video representations without full frame-by-frame supervision. Converting images into dynamic videos helps improve engagement and understanding compared to static depictions alone. As the coherence and resolution of image-to-video models continue improving, their potential applications in construction for bringing visuals to life will grow.

**Table 16:** Potential generative AI opportunities in the construction industry for image-video model type

| Potential opportunity | Description | Project phase |
|---|---|---|
| Design concept visualization | Generate animated walkthrough visualizations of architectural concept designs from still images. | Pre-construction |
| Site planning simulation | Produce simulated timelapse videos of site planning and layout from aerial photos. | Pre-construction |
| On-site safety analysis | Assess images of unsafe conditions and generate simulated incident recreations for safety analysis. | Construction |
| Construction progress | Compile timelapse videos of construction progress from daily site images. | Construction |
| Operation training videos | Produce equipment maintenance and operation training videos from instruction manual images and diagrams. | Post-construction |
| Promotional videos | Automatically generate engaging facility experience videos from images for leasing/sales. | Post-construction |
| Documentary videos | Compile construction progress, milestones, interviews, etc., into documentary-style recap videos from images. | All |

4.2.7. Video to text

AI transcription of video-to-text revolutionizes the construction industry by providing comprehensive documentation throughout the construction process. Potential applications of video-to-text conversion are shown in **Table 18.** During pre-construction, generative models could auto-transcribe kickoff meetings and regulatory compliance tutorial videos into concise text records. Safety briefing videos and daily progress meeting discussions on active construction sites could be translated into text summaries for distribution to wider stakeholders [104]. Post-construction commissioning and inspection videos also contain valuable verbal feedback that video-to-text techniques can structure into reports. Across phases, comprehensively transcribing archived project videos into indexed, searchable documentation enables robust retrospective analysis [105]. Models such as VideoCoCa can transcribe technical construction multimedia while filtering out irrelevant background noise [100]. The text outputs synthesize the key details and language from videos without needing to review hours of footage. This allows scaling extraction of vital audio information in rich multimedia that construction teams continuously generate. When deployed with proper data controls, video-to-text AI can unlock new levels of value from archived construction data without demanding extensive manual effort.

**Table 17:** Potential generative AI opportunities in the construction industry for video-text model type

| Potential opportunity | Description | Project phase |
|---|---|---|
| Meeting minutes generation | Automatically generate minutes from video recordings of project kickoff and planning meetings. | Pre-construction |
| Regulatory compliance briefs | Transcribe spoken content from regulatory compliance videos, creating textual briefs summarizing compliance requirements for the construction project. | Pre-construction |
| Safety briefing text summaries | Use generative AI to transcribe safety briefings in construction videos, generating concise text summaries for distribution to construction teams and stakeholders. | Construction |
| Defect detection | Analyze inspection videos and generate written alerts about potential issues for remediation. | Construction |
| Daily progress meeting transcripts | Transcribe discussions from daily progress meetings captured in videos, creating textual records of construction progress, challenges, and decisions. | Construction |
| Commissioning reports | Generate performance reports by transcribing functional testing/acceptance videos. | Post-construction |
| Project Archive | Create searchable records of the whole project by transcribing videos into indexed documentation. | All |

4.2.8. Video to image

As shown in **Table 19**, extracting key representative images from construction videos offers value across construction projects. Exploring site videos could be condensed into salient snapshots during pre-construction to accelerate assessments. Video conferences discussing design concepts can be automatically packed into a collage of snapshot visuals [108]. In the construction phase, delivery footage could be processed into consolidated photo logs of materials arriving on-site. Aerial construction video can generate periodic bird's-eye progress views [106]. Post-construction applications include extracting instructional stills from facility tutorials. Compiling time-lapse visual collages from archival videos can summarize entire project journeys. Video summarization techniques such as recurrent auto-encoders demonstrate capabilities to identify important frames that distill key visual concepts from longer video sequences [109]. Through the application of these techniques to construction footage, essential moments can be extracted without the necessity for manual video scrubbing. The representative thumbnail images could support rapid video review and summarization. They also integrate more seamlessly into reports and presentations compared to video embeds. Further innovation in dense video understanding and summarization will continue expanding the capabilities for automating the extraction of impactful visuals from construction multimedia.

**Table 18:** Potential generative AI opportunities in the construction industry for video-image model type

| Potential opportunity | Description | Project phase |
| --- | --- | --- |
| Site exploration frame extraction | Extract key frames from site exploration videos, creating static images that capture crucial moments and details for initial site assessments. | Pre-construction |
| Conceptual design snapshot generation | Extract representative snapshots from videos discussing conceptual designs and creating visual representations of architectural concepts for documentation and presentations. | Pre-construction |
| Virtual landscape preview stills | Generate still images from videos showcasing virtual landscape previews, providing stakeholders with static visual references for pre-construction landscape assessments. | Pre-construction |
| Material delivery visual logs | Generate representative photo logs from videos capturing materials and equipment as they arrive on site. | Construction |
| Remote and automated progress monitoring | Aerial video can be processed to automatically produce interval imagery depicting bird's-eye views of the site at different points in time. | Construction |
| Facility usage instructions still | Extract still images from videos providing facility usage instructions, creating visual stills that convey important guidelines for post-construction occupants. | Post-construction |
| Comprehensive project timeline collage | Compile critical frames from videos across all project phases into a comprehensive timeline collage, visually summarizing the entire project journey. | All |

4.2.9. Video to video

While static outputs enable analysis, video can engage stakeholders through dynamic visualization. Constructing the future requires envisioning it in motion. Generative video-to-video models can help bring these visions to life across the project lifecycle, as depicted in **Table 20**. During pre-construction planning and bidding, generative models could synthesize simulated construction sequences from source videos to allow interactive visualization of various work strategies and schedules for optimization [110]. On construction sites, input training videos could be adapted into multi-lingual versions translated across diverse crews to increase accessibility and comprehension. Architectural visualization videos, as-built documentation, and sensor data could be synthesized into lifecycle simulations and digital twin representations to support operations and maintenance. Opportunities exist across all phases, such as video quality enhancement, accelerated time-lapses, and video summarization for efficient review.

By learning spatial-temporal relationships from construction footage, models can extend source videos into modified outputs adapted for downstream requirements. This allows tailoring visual media for specific applications ranging from training to monitoring to forecasting. As video generation techniques continue advancing, the potential for AI-assisted video remixing and synthesis to enhance multimedia value across the construction project lifecycle will grow substantially [97]. Processing datasets accumulating from the proliferation of construction cameras and sensors using generative video models promises to unlock new visual insights and perspectives.

**Table 19:** Potential generative AI opportunities in the construction industry for video-video model type

| Potential opportunity | Description | Project phase |
|---|---|---|
| Cost Estimation | Synthesize timelapse estimates of different construction methods/schedules from sourced videos. | Pre-construction |
| User Experience Preview | By generating composite videos from design concept footage, animations, and 3D models, the expected user experiences and functional flows within a proposed development can be simulated before construction. | Pre-construction |
| Safety Orientations | Simulate hazard scenarios for training by compositing archived incident videos. | Construction |
| Multilingual Translation | Generate multilingual versions of instructional videos to support diverse crews. | Construction |
| On-Boarding Video | Produce guided video facility tours from archival documentation for onboarding and handoff. | Post-construction |
| Lifecycle Forecasting | Taking as-built construction videos detailing "as constructed" conditions, future renovations, retrofits, refurbishments, or capital upgrades planned at different phases of the asset lifespan can be digitally prototyped and overlaid. | Post-construction |
| Video Quality Enhancement | The upscale resolution, framerate, colors, etc., for legacy or damaged construction footage. | All |
| Accelerated Playback | Generate compressed timelapse videos from lengthy footage for rapid review. | All |

## 4.3. Challenges

The adoption of generative AI in the construction industry is growing; thus, it comes with challenges. As presented by the experts during the Delphi survey, these challenges are multi-faceted, encompassing domain-specific, technological, adoption, and ethical challenges, as shown in Figure 8. These challenges should be navigated for successful real-world deployment of generative AI in construction. A holistic approach considering technical and non-technical factors is required to overcome barriers and unlock the full potential of LLMs in construction.

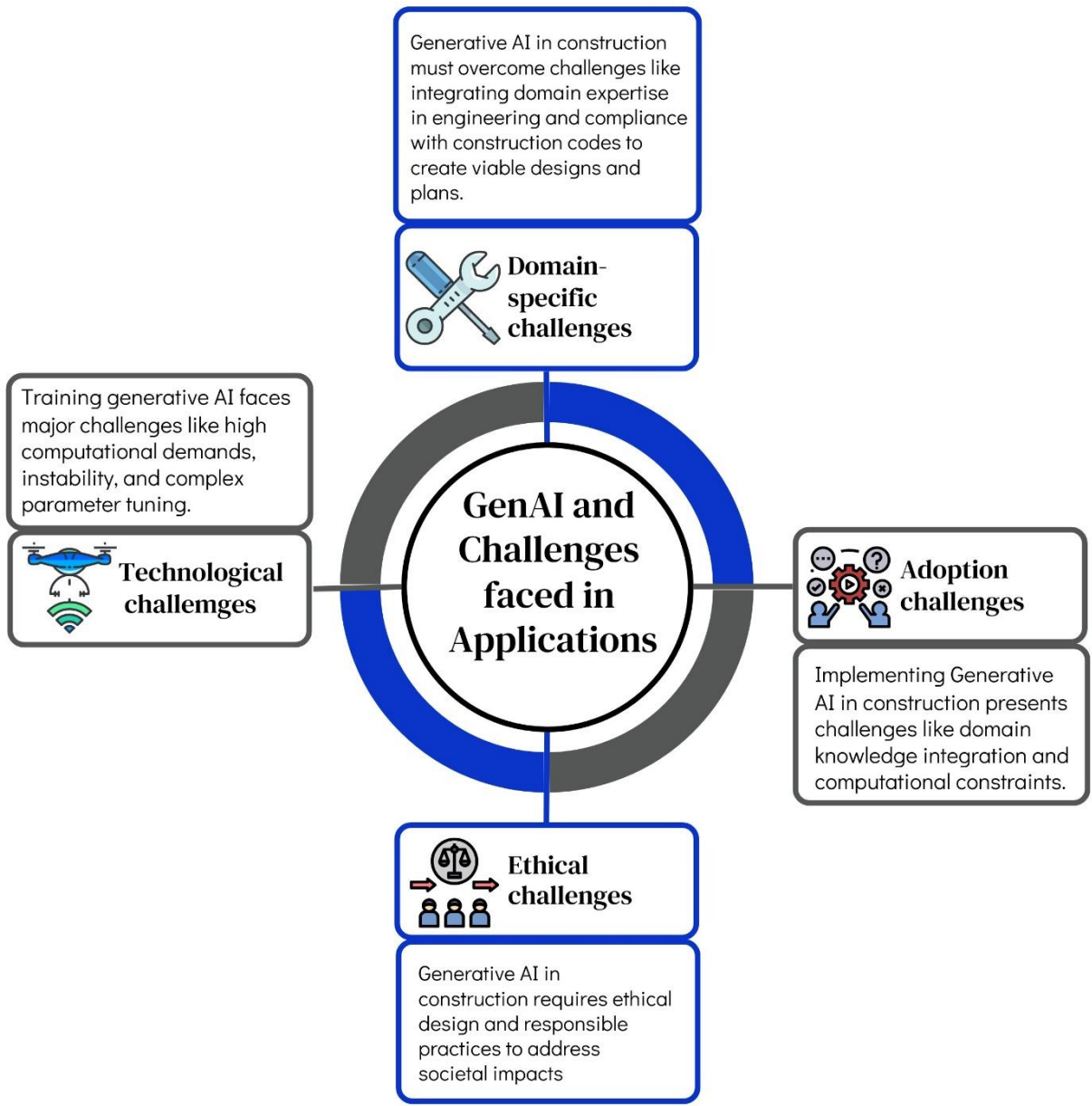

Figure 8. Common challenges faced in the application of GenAI in construction

4.3.1. Domain-specific challenges

*i. Requirement for construction-specific knowledge*

Construction is a complex field with intricate technical knowledge required to execute projects safely and efficiently [111]. However, most current generative AI techniques rely solely on statistical patterns extracted from data. They need the ability to explicitly encode the nuanced human expertise and domain constraints around structural engineering, materials science, construction codes, aesthetics, machinery, schedules, costs, etc. As a result, generative models trained exclusively on construction data may fail to produce valid, high-quality outputs that align with industry best practices. For example, a generative design model may create a visually appealing 3D building model that violates important structural principles, safety factors, or zoning regulations. The lack of engineering heuristics and constraints leads the unrestrained model to hallucinate flawed plans. Likewise, a generative text model trained only on construction documents will fail to generate specifications or instructions demonstrating a human's comprehension of materials compatibilities, sequences of operations, cost impacts, or equipment capabilities. Generative models need better integration of structured domain knowledge beyond just data patterns to reach their potential in construction. This remains challenging as industry experts' rules and mental models are difficult to codify for machines. Advances in neuro-symbolic AI, modular architectures, and expert-in-the-loop training show promise for imbuing models with more robust construction domain intelligence [112].

ii. *Handling unstructured and heterogeneous data*

Construction data exists in multifaceted, unorganized formats across disparate systems, posing challenges for generative AI. Project information encompasses everything from scanned paper blueprints to 3D BIM models, permits, contracts, change orders, requests for information (RFIs), submittals, specifications, budgets, meeting minutes, multimedia, and more. These data types have different structures, semantics, units, symbols, file formats, and modalities. Generative models like GANs and VAEs struggle to ingest this heterogeneous, unstructured data directly to synthesize coherent outputs. For example, a basic image-to-image model cannot map a 3D BIM model, change order form, and permit application into a unified generated output. The variability across projects also hampers standardized tooling. Each construction firm may have customized data conventions, nomenclatures, templates, and systems tailored to their needs. Creating consolidated datasets from dispersed historical records is arduous. To work around these challenges, purpose-built multi-modal generative architectures are necessary [113]. Techniques like attention mechanisms, graph networks, and transformer models show promise for learning alignments and correlations across varied data types. However, no universal solution exists to handle the messiness of real-world construction data. Generative AI still requires extensive wrangling of unstructured inputs into tidy, normalized features.

iii. *Lack of large-curated datasets*

Generative AI models need massive, high-quality training datasets to perform well. However, most construction firms do not consistently organize and consolidate their project data into formats usable for training a large model [103]. Historical records remain fragmented across various databases, file shares, and systems. The effort required to aggregate and clean unstructured construction data into coherent datasets is prohibitive without dedicated workflows. Data may reside in legacy formats. Important contextual links between related data points may be lost. The need for more data versioning, consistency, and curation poses challenges. For niche construction applications like generating site layouts or drywall specifications, virtually no large canonical datasets exist publicly to train models [114]. Collecting sufficient data from scratch requires substantial industry participation across firms. Annotation and labeling also necessitate scarce expert time. Without sizable, high-coverage training datasets, generative models struggle to generalize. They easily neglect sparse edge cases or unique scenarios found in complex construction projects. Models trained on inadequate data produce lower-fidelity outputs that lack realism and conformity to standards. Overcoming this bottleneck will require construction firms to systematically organize data accumulation, annotation, versioning, and consolidation workflows. Precompetitive industry data consortiums can also help aggregate datasets for typical AI applications.

*iv.    Bias in existing datasets*

Construction datasets often exhibit significant regional biases propagated through generative AI models trained on this data. For example, architectural plans and building methods reflect local materials availability, weather patterns, seismic requirements, and zoning laws. Specifications follow jurisdiction-specific codes and standards. Units of measure, terminology, and language also vary geographically [115]. If models are trained solely on historical datasets from a particular country or city, the generated outputs will inherit these narrow perspectives. A design model trained only on American examples may overlook important considerations for cyclone-prone regions when deployed in the tropics. Likewise, language models trained only on specifications for a particular state could generate confusing RFI responses for international contractors following different norms. Outputs may also inadvertently include inapplicable regional colloquialisms. Training datasets must include wide diversity along multiple geographic axes to minimize bias to improve model coverage. However, thoughtfully collecting and curating such datasets is challenging for firms focused on their local region. Synthetic data augmentation techniques can help artificially expand variety once baseline data is available [116]. In practice, biased training sets often necessitate maintaining individualized models tailored to each application region. But this multiplicity hampers scaling and adds overhead. Developing adaptable generative models that generalize across diverse contexts remains an impactful challenge in construction.

    v.    *Integration with workflows and standards*

The construction industry has relied heavily on workflows and proprietary systems tailored to each firm's needs and project requirements for many years. Seamlessly integrating generative AI solutions with these incumbent environments poses significant adoption difficulties. A core challenge is the need for interoperability between the modern ML tools underpinning generative models and the fragmented legacy software prevalent in construction. Custom integrations are needed to connect predictive models with databases, analytics dashboards, enterprise resource planning (ERP) platforms, BIM tools, and more [103]. However, construction systems often lack application programming interfaces (APIs). Generative models also need flexibility to adapt outputs to the proprietary data structures, nomenclatures, and templates used within each company. One-size-fits-all solutions struggle without customization. Firms are also reluctant to overhaul proven workflows solely to accommodate AI systems that appear disconnected from daily tasks. For adoption, generative models should directly build on available in-house data while aligning outputs to industry-standard specifications, equipment libraries, materials databases, regulations, and best practices. Workers are more inclined to use AI content that meshes with familiar domain paradigms rather than introducing foreign concepts. Overcoming these integration hurdles requires either extensive custom development efforts or architectures adaptable enough to map generative outputs to diverse construction environments out of the box. Finding the right balance between generalization and specificity remains an obstacle to embedding AI within incumbent workflows.

4.3.2. Technological challenges

    i.    *Model instability and training difficulties*

Training generative AI models like GANs, GPTs, and LLMs to reliably produce stable, high-quality outputs remains challenging, even in broader application areas outside construction. [117]. These training and stability issues become even more pronounced in the complex, constrained construction industry. The non-linear neural network architectures underlying many generative models have billions of parameters optimized through stochastic gradient descent [33]. The internal representations and dynamics of these massive models still need to be better understood, making their unpredictability harder to troubleshoot. During training, generative models are prone to problems like mode collapse, failing to capture the full diversity of training data. Finding the right balance between overfitting the data while still being able to generalize is tricky. Other issues, like vanishing gradients, can prevent networks from adequately learning. These training instabilities are amplified when models are scaled to handle sizeable multi-modal construction datasets. Getting models to synthesize completely novel outputs unrestrained by training patterns, as required in generative tasks, also increases unpredictability. Advances in principled network design, normalization techniques, robust optimization algorithms, and better training diagnostics should improve model stability. But for now, the opacity and fragility of uncontrolled generative synthesis pose an inherent challenge.

ii. *Computational resource requirements*

Generative AI models are extremely computationally intensive, both during training and inference. State-of-the-art models like GPT-4 contain billions of parameters, requiring extensive parallel processing power on specialized hardware like GPU clusters or TPUs to train within reasonable timeframes [103]. For smaller construction firms, procuring and operating this expensive infrastructure may be infeasible just for experimenting with generative AI. Outsourcing to cloud platforms can mitigate costs but still demands significant investment. The carbon emissions footprint from model training should also be considered, given sustainability goals in the industry [118]. Even after models are trained, deploying them for inference and generating new outputs is resource-intensive. Real-time generation of high-resolution images, 3D models, or lengthy text would require low-latency access to powerful cloud computing. Many construction companies need more modern on-demand computing resources. As model sizes and demand for higher-quality outputs increase, so will hardware requirements. Construction firms without the IT infrastructure or budgets to continuously upgrade generative AI capabilities risk being left behind. This could create a bifurcation where only the most prominent players can afford to operate at the state-of-the-art. Advancement of more efficient architecture, distillation techniques, and on-device inference chips may eventually dampen costs. But in the interim, the level of resources needed to benefit from generative AI poses barriers, especially for smaller general contractors and subcontractors. Strategic partnerships with tech providers could help navigate the substantial computing investments involved.

iii. *Assessing output quality*

Unlike discriminative ML models, where accuracy metrics quantify performance, evaluating generative AI outputs' true quality is difficult. Metrics like Fréchet Inception Distance provide a proxy for similarity to accurate data distributions. However, these have limited utility when outputs are meant to be completely novel syntheses tapping the unknown. For niche construction applications, benchmarking datasets to test against do not exist. Assessing quality often relies on slow and subjective human review by domain experts, which does not scale. Furthermore, generated outputs like text, images, or 3D models may appear convincing on the surface to non-experts, exhibiting clear style and coherence [119]. However, upon closer expert inspection, these outputs lack deep domain-specific fidelity and violate constraints that may be obvious to a construction professional. Detecting these subtle faults, which do not manifest in surface metrics, remains an open problem. Developing and integrating better quality assurance techniques for generative AI in construction is crucial. This will likely require a combination of automated quantitative checks, qualification processes, and skilled human reviewers. Without rigorous validation protocols, using generative models for safety and cost-critical construction tasks is precarious. All stakeholders need reliable indicators that system outputs meet domain requirements before fully embracing generative techniques.

*iv.  Potential for hallucination and factual inconsistencies*

A significant danger posed by generative AI is its tendency to fabricate imaginary details that appear valid but diverge from reality[30]. When synthesizing novel outputs, these models are unrestrained by the fixed training data distribution. The systems "hallucinate" new content by stochastically combining learned features and patterns. In open domains like art and entertainment, such an unconstrained generation of new ideas may be desired. But for the safety-critical construction industry, factual inconsistencies or false details could have disastrous consequences if relied upon. Even minute defects in a generated building design, equipment specification, or work procedure could lead to accidents, delays, or rework down the line. Unlike discriminative models, which stick tightly to input features, generative models have free rein to distort outputs during synthesis. While coherence and surface plausibility remain high, factual correctness often suffers [103]. Without proper oversight, these distortions go unnoticed until problems arise in construction or operations. The unpredictable, unsupervised nature of generative AI makes it fundamentally risky for domains requiring tight conformance like construction. Extensive validation processes led by human experts and automated safety checks are necessary when applying these models. However, detecting the subtle faults unique to generative approaches remains an open research problem. Only when more controlled techniques are developed, unleashing unconstrained generative models comes with high uncertainty. Their propensity to smoothly fabricate imaginary details outside the training distribution should instill caution. While promising, balancing generative AI's creative potential with construction constraints is critical.

*v.   Lack of explainability*

A significant limitation of modern generative AI techniques is their black-box nature. Systems like LLMs and GANs offer minimal transparency into their internal reasoning for producing specific outputs over others [120]. The models synthesize outputs by propagating input signals through billions of transformations across neural network layers. Explaining why one output manifested versus another is nearly impossible given this complexity. In construction, lack of explainability poses risks and makes diagnosing errors harder. When designs, images, or text are generated, professionals have no visibility into the generative model's intent or rationale. This needs to be revised in order to maintain human oversight of the system's thinking and conclusions. If flaws are detected, the opaque models provide little clue into the root causes. Troubleshooting and correcting errors becomes a guessing game without explanatory abilities. This could lead to blind trial-and-error tuning versus informed debugging. More transparent and controllable architectures may be needed for broader acceptance in the relatively conservative construction industry. Hybrid approaches combining neural networks with declarative knowledge about engineering constraints could improve interpretability. Interactive interfaces that allow step-by-step manipulation of generative models also offer more transparency.

### 4.3.3   Adoption challenges

*i.   Resistance to new technologies*

The construction industry has historically needed to be faster to adopt new technologies compared to other sectors. This inertia and resistance to change stems from several interrelated factors. Many construction firms rely heavily on established processes and workflows that have been incrementally optimized over the decades. There is often a reluctance to modify or replace these proven legacy, deeply ingrained methods [121]. Furthermore, the supply chain involves disparate stakeholders with different capabilities and resources. Aligning on new technology adoption is difficult across this fragmented ecosystem. At a management level, there are concerns that AI could disrupt traditional roles and ways of doing business in construction. The industry relies on specialized trades and processes that workers have invested years into mastering. Introducing unfamiliar systems feels inherently risky, making management hesitant to champion large-scale technology overhauls. Overcoming these barriers will require a combination of peer-based advocacy, demonstrable benefits, incentives, change management planning, and strong leadership buy-in.

## ii. Lack of skills and expertise

The use of generative AI requires specialized skills that currently need to be improved in most construction companies [30]. While these firms have deep domain expertise in construction processes, materials, equipment, etc., they have limited in-house experience with AI and data science. Most construction companies cannot realistically build large internal AI teams from scratch. Construction firms will likely need to hire dedicated AI talent or partner with technology firms to complement their domain knowledge. However, professionals with deep AI expertise and construction industry knowledge are rare and difficult to recruit. Closing the skills gap will require a combination of recruitment, training, partnerships, and creating more no-code or low-code solutions tailored to the industry.

## iii. High upfront investment costs

Adopting generative AI poses considerable upfront investment costs, which may deter construction firms from pursuing it. Firstly, data preparation requires aggregating dispersed historical data from multiple systems and getting it into a unified format [122]. Next, licensing and developing generative models necessitates paying for specialized AI services. The computational resources needed for training and inference, such as cloud GPUs, add to the technology bill. Integrating the AI system with existing construction workflows and IT infrastructure demands custom development efforts [103]. Finally, machine learning engineers incur ongoing maintenance, monitoring, and enhancement costs. For large construction corporations, these expenses may be feasible to absorb. However, smaller contractors and trade firms operate on tighter margins and budgets. Many may find the capital expenditures required to implement generative AI prohibitively high. The construction industry needs to be more fast-moving initiatives on investments, especially for emerging technologies like generative AI. Demonstrating a convincing return on investment is critical for securing buy-in.

## iv. Immature supporting infrastructure

Successfully implementing generative AI requires data infrastructure and workflows, which are currently immature in the construction industry. Firstly, most firms lack the data pipelines and consolidation needed to feed massive training datasets to generative models. Data labeling and annotation workflows necessary for supervision are also non-existent. Furthermore, the machine learning operations (MLOps) tools for versioning models, monitoring systems, and ongoing improvement are foreign to most construction IT departments. Generative AI relies extensively on the scale of computing power, demanding integration of construction data systems with cloud platforms [14]. However, seamless connections between internal databases, BIM models, and external cloud resources are rare [123]. There is also a shortage of prebuilt integrations between construction software tools and generative AI APIs. The surrounding ecosystem to enable enterprise adoption is still evolving. In effect, construction firms cannot simply plug and play off-the-shelf generative AI solutions into their existing IT systems. Substantial infrastructure development and integration efforts are required to create the data and compute foundations. For many companies, this necessitates a complete overhaul of internal data practices, development stacks, and system architectures.

  v. *Unclear governance frameworks*

There are unresolved questions around legal liability - who is accountable if an AI system produces faulty designs, specifications, or recommendations that lead to accidents? Quality control and validation protocols for generative models in construction are also lacking. Furthermore, the security implications of relying on AI to guide mission-critical construction processes are still being worked out. Risk management frameworks and technical standards have not caught up to the rapid advances in generative techniques. There are also ethical concerns about reproducing historical biases in data, which require governance to be addressed transparently and responsibly. The regulatory regime surrounding generative AI in construction is unclear and fragmented. Companies are hesitant to deploy unproven technologies without best practices or precedents to follow. Both public and private institutions need clear legal guidelines, technical validation protocols, model risk management expectations, and standards of use [124].

4.3.4  Ethical challenges
  i. *Data privacy and security*

Construction projects generate vast amounts of potentially sensitive data - from financial records to design specifications to site photographs. As this data is increasingly used to train generative AI models, firms must act responsibly to respect privacy and maintain trust [122]. However, most construction data practices are focused on operations rather than ML readiness.

Efforts will be needed to obtain proper consent, audit datasets, and implement access controls for AI systems. Data anonymization techniques can help remove personally identifiable information. But details like project names, locations, and dates often cannot be fully stripped without losing utility. Strict governance models for internal data collection, external usage, and retention will need to be developed. Cybersecurity is also critical, given the highly sensitive nature of commercial construction data. Breaches during model development or deployment could have serious consequences ranging from confidentiality violations to industrial espionage. Construction firms can uphold privacy while tapping AI advancements by minimizing risks through responsible data curation, anonymization where possible, and tight access restrictions. However, this may require overhauling ingrained data practices focused on operational efficiency and ethics. The cultural and procedural shifts will challenge organizations to harmonize AI progress with core principles of trust and transparency [125].

*ii.  Social concerns about job automation*

With the potential to enhance many human tasks, the adoption of generative AI in construction raises understandable concerns about workforce impacts. However, the effects are unlikely to be straightforward substitution of workers. AI may automate narrow, repetitive tasks but augment professionals to be more productive on complex strategic initiatives [126]. New human roles overseeing and collaborating with AI systems will also emerge. Proactive communication, training programs, and organizational change management will be imperative for a responsible transition. Leaders must be cognizant of apprehensions among workers fearing replacement by "thinking machines." Construction firms that are reliant on specialized trades have a particular responsibility to involve and support affected staff through an AI-enabled transformation. Instead of blunt displacement, AI should aim for symbiosis - enhancing professionals' capabilities while handling rote work. Adoption with the right intention of uplifting workers and augmenting expertise can help construction firms achieve societal benefits and sustainable competitiveness.

*iii.  Potential for misuse*

The autonomous and scalable capabilities of generative models create risks of misuse if deployed irresponsibly. For instance, AI systems lacking appropriate safeguards could generate realistic but structurally flawed building or equipment designs. Without rigorous engineering constraints and oversight, the unrestrained creativity of generative models could produce designs that circumvent safety codes and regulations [119]. Similarly, project plans, budgets, certificates, invoices, change orders, and other documentation falsified by AI could enable fraud or errors. The ease of generating convincing paperwork at scale for malicious purposes poses financial and legal risks. To prevent misuse, construction firms need to implement extensive technical and ethical precautions [127]. This includes carefully auditing training data and models for issues like bias, establishing sandboxed development environments, verifying outputs, and instituting human-in-the-loop checks before deployment. Responsible governance encompassing explainability, transparency, and accountability is also critical. Generative AI offers immense opportunities but also risks if its capabilities are unleashed carelessly. With prudent controls and oversight, construction professionals can minimize hazards while benefiting from accelerated innovation.

# 5. Framework for building custom LGM in the construction industry

While general pre-trained models like GPT-4 and Gemini Pro offer promising capabilities [45,95], developing LGMs customized for the construction domain can further enhance performance on industry-specific tasks. This section provides a framework that construction professionals and firms can follow to develop tailored LGMs using their proprietary data (see Figure 3). The key steps include construction data collection, dataset curation, training of custom LGM, evaluation of the LGM, and deployment.

## 5.1. Construction data collection

The first critical step in developing a custom LGM for construction is aggregating a broad corpus of relevant data from past projects and documentation across the firm. This serves as the foundation for training the model to comprehend and generate high-quality industry-specific language [103]. The data should be pulled from a diverse range of historical sources to cover the full breadth of concepts and terminology used within the company's work. Potential sources that should be tapped include technical specifications, equipment manuals, permit applications, contractor invoices, design reports, construction schedules, requests for information, project budgets, safety protocols, inspection checklists, as-built drawings and videos, relevant codes and standards, project contracts, and meeting minutes [86]. Essentially, all unstructured data around projects both directly produced by the firm and exchanged with partners, contains valuable language samples that can educate the LGM. Ideally, the data collection should draw from both successful and problematic construction projects within the company's archives. This provides balanced examples and helps the model better handle edge cases by learning from challenging historical incidents. Maximum diversity in the kinds of projects covered also allows the LGM to generalize robustly. In terms of format, the text data should be structured into machine-readable JSON if readily available in this format within the firm's document systems [26]. However, extensive cleaning and preprocessing of diverse unstructured data will likely be required. For scanned or image-based data, optical character recognition can extract text. For legacy video and audio, speech recognition techniques can generate transcripts. Metadata extraction can pull useful tags and descriptions from media files. Point cloud data may need processing into voxel grids or meshes. The raw data extracted across modalities like text, image, video, and audio must then be transformed into standardized corpora in formats digestible for model training. Expect substantial effort to munge multifaceted data sources into shapes consumable by generative algorithms.

### 5.2. Dataset curation

After aggregating raw data, the next step is carefully curating it into a high-quality dataset ready for training the construction LGM. This involves extensive processing and analysis. First, any sensitive personal information or proprietary business data must be removed from the corpus to respect privacy and security protocols. Next, identifiable entities like specific project names and locations should be anonymized where possible to mitigate risks. Thorough cleaning is required to fix any formatting inconsistencies, OCR errors, or annotation issues so that the data is pristine. The data sources should also be analyzed to ensure sufficient diversity - if the dataset focuses too heavily on certain project types or documentation formats, it can lead to a lack of broad applicability. Chronological splitting into train, validation, and test sets is also critical for properly evaluating model performance over time [15,32]. Moreover, synthesizing additional diverse examples through techniques like contextual data augmentation should be considered to boost the coverage of niche cases. Domain experts should manually sanity-check random samples from across the final dataset to catch any lingering issues before training begins. This human-in-the-loop auditing step provides quality control and ensures the data is aligned with true construction language [128].

### 5.3. Training custom LGM

With a tailored construction-specific dataset prepared through careful data curation, the next phase is leveraging this data to train a custom LGM for construction tasks. The model can be initialized from scratch and trained end-to-end on just the domain data. However, it is also common to initialize with an existing general pre-trained LGM like GPT-4 that has already learned strong linguistic representations from web-scale corpora [45]. This foundation can then be fine-tuned on the construction dataset to adapt to industry-specific terminology and patterns. Transfer learning in this manner can significantly reduce the computational resources and time required for training versus a from-scratch approach [87]. Regardless of the initialization technique, the overall training methodology involves first selecting an appropriate underlying model architecture and size. Transformer networks currently demonstrate state-of-the-art performance on language tasks but require tuning of their complex configurations to fit each dataset and use case [46]. Training of the model is then conducted using GPU or TPU computational infrastructure until convergence on the construction training data distribution as measured by validation set performance. The training hyperparameters, including batch size, learning rate schedules, and activation functions, must be finely tuned based on the validation results to optimize model quality.

This entire model development process requires extensive computational resources for both the core training and the surrounding hyperparameter optimization. An opportunity to enhance the training using retrieval augmented generation (RAG) also exists [129]. A retrieval system can be built on top of the construction dataset to provide contextual examples that keep the LGM grounded in domain-specific language during training. Hence, developing a high-quality, performant LGM customized to construction data involves synthesizing diverse techniques into a robust training methodology.

## 5.4. LGM evaluation

Once the LLM is trained on construction data, it must undergo rigorous evaluation before deployment to ensure it achieves the performance and quality thresholds required in downstream applications. The semantic coherence, grammar, terminology, and validity of generated outputs should be extensively assessed via qualitative human review by domain experts. This allows for validating that the model produces high-quality language aligned with true construction concepts. Checking for potential biases and factual inaccuracies is also critical to avoid operational risks. In addition, the model should be quantitatively benchmarked against baseline methods on domain-specific tasks using relevant metrics. For instance, the customized LGM can be evaluated for construction project phase classification accuracy and compared to off-the-shelf generic models. Other quantitative tests might include cross-referencing generated project budgets against actual data to assess fidelity [130]. Any shortcomings identified during evaluation should be addressed by re-training the model using modified data that improves coverage of deficient areas or adjusting the model architecture itself. The evaluation process also provides feedback for additional training to continue enhancing the LGM post-deployment.

## 5.5. Custom LGM deployment

To enable scalable deployment, the model should first be containerized using technologies like Docker and Kubernetes [131]. This encapsulates the model in a portable package with libraries and dependencies while managing computational resources. Exposing a high-performance API or web interface allows sending inference requests to the containerized model server. This powers integration into downstream domain applications. For instance, the custom LGM could be embedded within AI assistants, document generators, project recommendation systems, and other construction tools that benefit from its specialized text generation capabilities. Robust MLOps processes need to be implemented for continuous monitoring, versioning, and improvement of the model post-deployment [132]. As new project data comes in, it can be used to further tune and enhance the model to stay up to date. Human oversight and governance are critical during deployment to ensure quality control and responsible privacy, security, and ethics practices. With the proper infrastructure and processes in place, the custom construction LGM can be sustainably integrated to augment a wide range of business functions with an industry-tailored AI generation solution. This framework provides a methodical blueprint for construction firms to transform their data into strategic generative capabilities.

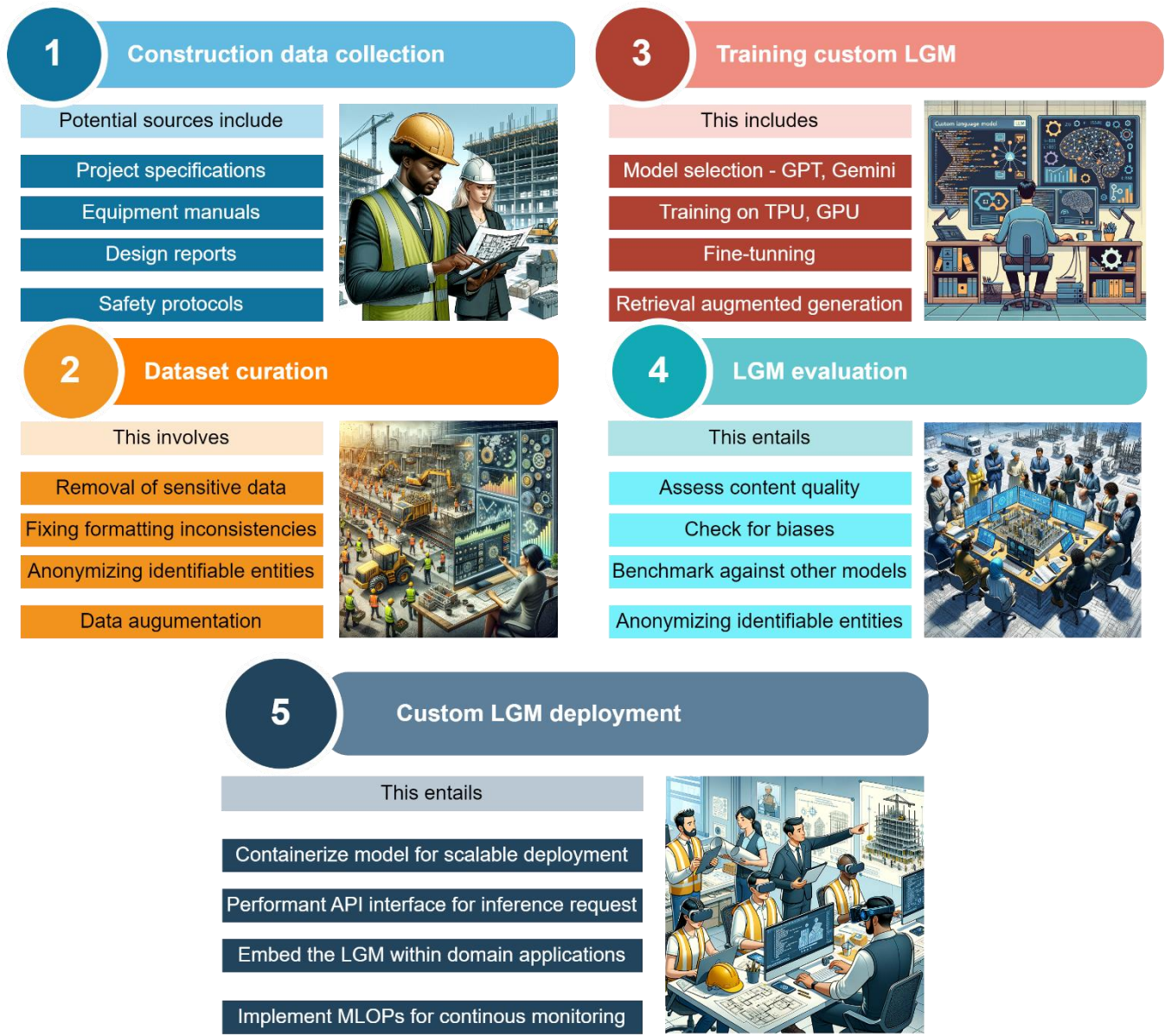

**Figure 9:** Framework for building custom LGM in the construction industry

## 6. Case study – information retrieval and knowledge discovery

To validate the potential for using generative AI in construction, a case study was conducted focused on information retrieval and knowledge discovery. This is one of the potential opportunities identified in the previous section on text-to-text applications. Querying contract documents is a valuable application, as contract documents contain critical project requirements and details but can be lengthy and complex to manually search through. The contract document employed for this case study was obtained from a consultancy firm that served as the project manager. The project entailed the construction of a three-story hostel facility for a higher education institution. The contract document contained key information on project scope, specifications, materials, timelines, costs, quality standards, and other crucial parameters. As contract documents like this are often dense and unstructured, retrieving information requires tedious manual searching. Generative AI offers the ability to query the document in natural language and receive direct answers summarizing the most relevant details. This case study demonstrates the value of training generative models on real-world construction contracts to improve information search and extraction.

## 6.1. Model development

For this case study, the GPT-4 model was leveraged as the base LLM. GPT-4 is a proprietary LLM designed by OpenAI to generate coherent and useful text in a wide variety of domains [95]. It was pretrained on massive text datasets encompassing diverse topics and demonstrated state-of-the-art natural language processing capabilities [133]. To further enhance the capabilities of GPT-4 for the construction contract domain, a RAG framework was implemented on top of the base model. RAG integrates semantic search over a domain-specific knowledge base into the text generation pipeline. This allows retrieving the most relevant contextual examples from the contract text to prime the LLM when responding to queries. The RAG framework helps ground the model's outputs in the actual contract content, avoiding hallucinations.

As depicted in Figure 10, the RAG pipeline consisted of [134]:

- **Importing the contract document**: The first step was ingesting the raw text data from the contract document into the RAG system. This included preprocessed documents in PDF format.
- **Splitting documents into coherent chunks**: The full contract document was segmented into smaller chunks of text spanning 3-5 sentences focused on a coherent part of the contract. This allowed more fine-grained contextual retrieval.
- **Creating embeddings for the chunks**: ML embeddings were generated using an advanced semantic encoding model for each text chunk. OpenAI embedding was used for this purpose via API access.
- **Storing the chunk embeddings in a vector knowledge base**: The chunk embeddings were indexed in a high-performance vector database. Cassandra database was used for this purpose [135]. This enables quick retrieval of contextually similar chunks.
- **Accepting user query as input**: At inference time, the user provides a text query expressing their desired contract document information need.

- **Embedding query into same vector space**: The input query is encoded into the same semantic vector space as the chunks using the same sentence model.
- **Performing semantic search to identify relevant specification chunks**: Efficient approximate nearest neighbor (ANN) search is run to find chunks with the highest semantic similarity to the query vector.
- **Ranking retrieved chunks by semantic similarity**: The topmost similar chunks are ranked and filtered to create a subset most relevant to the query.
- **Providing top k chunks to guide LLM text generation**: The top-ranking contract document chunks are provided as contextual examples to prime the LLM to generate focused and valid contract text.

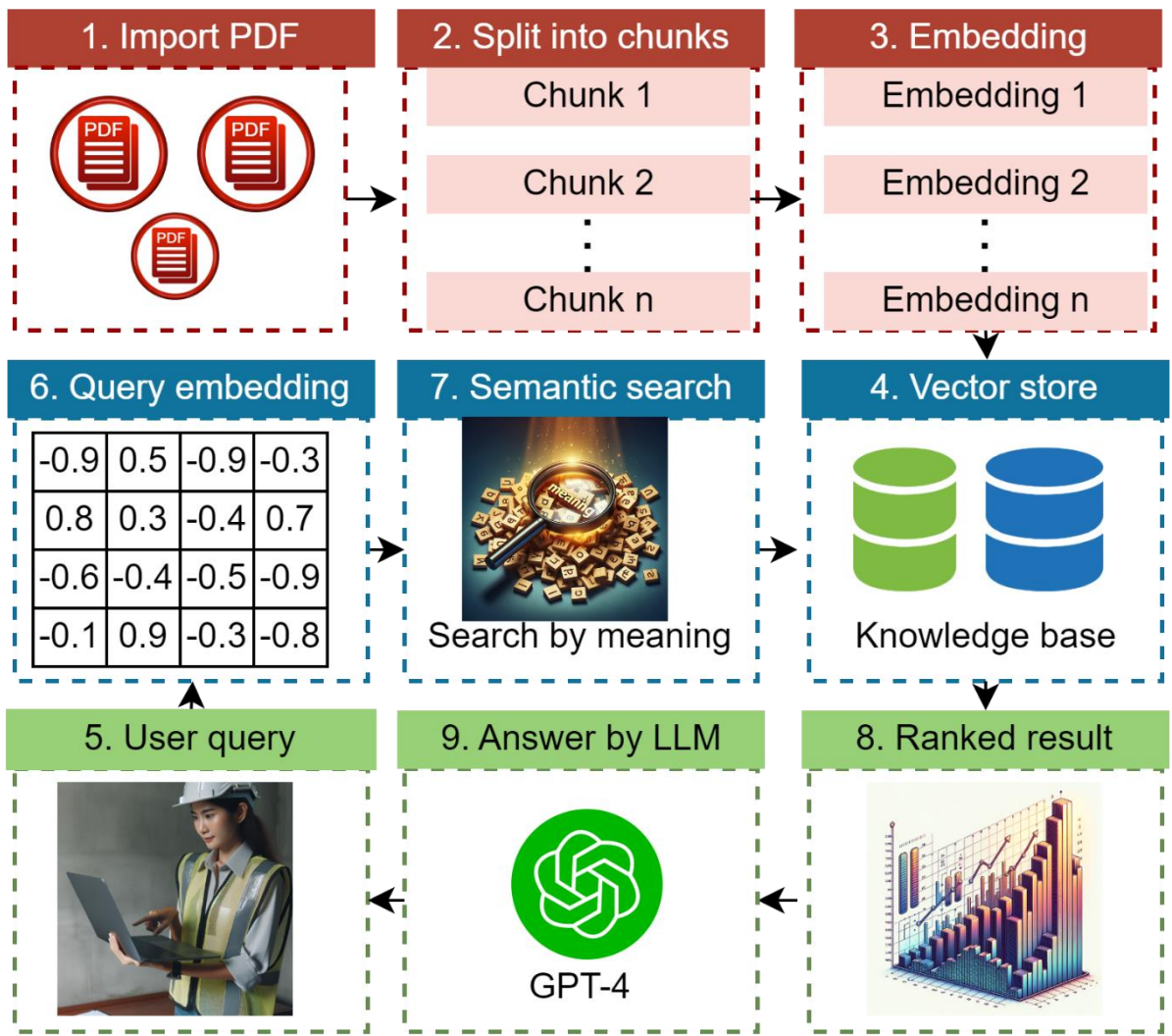

Figure 10: Retrieval augmented generation pipeline

## 6.2. Model evaluation

To develop an appropriate set of queries for evaluating the model, 30 potential natural language questions were initially formulated based on information contained within the contract document. These draft questions were reviewed by a panel of three experts with professional construction knowledge to validate that they represent realistic queries of interest to industry practitioners accessing such a document. The first expert confirmed 26 of the proposed questions as relevant, while the other two experts each validated 23 questions. By taking the intersection, a final set of 20 common questions validated by all three experts was derived. This cross-validated question set encompasses diverse query types covering key information needs that construction professionals would seek to retrieve from contract documents. The 20 expert-approved questions were employed to evaluate the models' performance at extracting relevant answers from the contract.

The approved questions and contract document were provided to 3 experts from the original panel that validated generative AI opportunities and challenges. These experts evaluated the model's responses to each question based on four metrics, which are similar to the metrics adopted by Wolfel et al. [136]:

- Answer - Assesses if the model provides a substantive response ("Yes") or a non-answer like "I don't know" ("No").
- Quality - Rates the truthfulness of the response on a 5-point scale.
- Relevance - Rates how relevant the response is to the query and contract on a 5-point scale.
- Reproducibility - Assesses the consistency of responses to the same question on a 5-point scale.

The average score (mode was employed for the "answer" metric) for each question was calculated across raters for both the baseline GPT-4 model and the GPT-4 plus RAG system. The results are shown in Table 21, enabling a quantitative comparison of the two models' performance in extracting accurate, relevant information from the contract document when queried in natural language. According to Table 21, the high answer rate of 100% for GPT-4 indicates it consistently provides substantive responses to the questions rather than failing to generate any reply. However, the lower quality score of 3.87 reveals some responses fabrication details are not actually present in the contract, as the model hallucinates plausible-sounding but incorrect information. The decent relevance rating of 4.01 shows GPT-4's outputs are topically on point but strained by invented content. The reproducibility score of 4.53 suggests some inconsistency across repeated queries as well. In comparison, the RAG-enhanced GPT-4 model achieves higher quality and relevance ratings of 4.13 and 4.48, demonstrating improved faithfulness through grounding outputs in retrieved contract passages. This reduces hallucinated content substantially. The superior reproducibility of 4.77 also highlights more stability from RAG's contextual retrieval. However, the lower 90% answer rate points to limitations in linking some questions to pertinent evidence, causing the model to default to "I don't know" non-answers. The quantitative metrics illustrate RAG's ability to enhance faithfulness and mitigate risks of hallucinations that generative models like GPT-4 exhibit.

**Table 20:** Custom LLM model evaluation

| Question Number | Model | Answer | Quality (1-5) | Relation (1-5) | Reproducibility (1-5) |
|---|---|---|---|---|---|
| 1 | GPT-4 | Yes | 5.00 | 5.00 | 5.00 |
| 2 | | Yes | 5.00 | 5.00 | 5.00 |
| 3 | | Yes | 1.00 | 1.33 | 5.00 |
| 4 | | Yes | 2.67 | 1.67 | 3.00 |
| 5 | | Yes | 1.33 | 2.00 | 5.00 |
| 6 | | Yes | 5.00 | 5.00 | 4.00 |
| 7 | | Yes | 5.00 | 5.00 | 5.00 |
| 8 | | Yes | 4.33 | 4.67 | 4.00 |
| 9 | | Yes | 5.00 | 4.00 | 3.67 |
| 10 | | Yes | 3.00 | 3.00 | 5.00 |
| 11 | | Yes | 5.00 | 5.00 | 5.00 |
| 12 | | Yes | 5.00 | 5.00 | 5.00 |
| 13 | | Yes | 4.33 | 5.00 | 5.00 |
| 14 | | Yes | 5.00 | 5.00 | 5.00 |
| 15 | | Yes | 1.67 | 4.00 | 5.00 |
| 16 | | Yes | 4.33 | 5.00 | 5.00 |
| 17 | | Yes | 5.00 | 5.00 | 5.00 |
| 18 | | Yes | 4.00 | 4.00 | 4.00 |
| 19 | | Yes | 1.67 | 1.67 | 2.00 |
| 20 | | Yes | 4.00 | 4.00 | 5.00 |
| Average (Percentage) | | 100% | 3.87 (77.4%) | 4.01 (80.2%) | 4.53 (90.6%) |
| | GPT 4 + RAG | | | | |
| 1 | | Yes | 4.00 | 5.00 | 5.00 |
| 2 | | Yes | 5.00 | 5.00 | 5.00 |
| 3 | | No | - | - | 5.00 |
| 4 | | Yes | 3.67 | 2.33 | 4.00 |
| 5 | | Yes | 5.00 | 5.00 | 5.00 |
| 6 | | Yes | 5.00 | 5.00 | 5.00 |
| 7 | | Yes | 5.00 | 5.00 | 5.00 |
| 8 | | Yes | 5.00 | 5.00 | 5.00 |
| 9 | | Yes | 4.33 | 5.00 | 4.67 |
| 10 | | Yes | 2.33 | 4.00 | 2.00 |
| 11 | | Yes | 1.67 | 5.00 | 4.67 |
| 12 | | Yes | 5.00 | 5.00 | 5.00 |
| 13 | | Yes | 4.67 | 4.33 | 5.00 |
| 14 | | Yes | 4.00 | 5.00 | 5.00 |
| 15 | | Yes | 4.00 | 5.00 | 5.00 |
| 16 | | Yes | 1.67 | 1.33 | 5.00 |
| 17 | | Yes | 4.00 | 5.00 | 5.00 |
| 18 | | Yes | 5.00 | 3.67 | 5.00 |
| 19 | | No | - | - | 5.00 |
| 20 | | Yes | 5.00 | 5.00 | 5.00 |
| Average (Percentage) | | 90% | 4.13 (82.6%) | 4.48 (89.6%) | 4.77 (95.4%) |

Figure 11 provides example query-response screenshots for questions 6 and 15, comparing the baseline GPT-4 model and the GPT-4 plus RAG system. Figures 11a and 11d show the GPT-4 responses to these questions, while Figures 11b and 11e display the responses augmented by the RAG retrieval pipeline. Figures 11c and 11f highlight the relevant passages containing the answers in the original contract document. Examination of the examples illustrates that both GPT-4 and GPT-4 + RAG correctly answered question 6. However, for question 15, GPT-4 hallucinates details about GCC Clause 44 that are not contained in the actual contract, while GPT-4 + RAG grounds its response in the original document context to extract the price adjustment formula accurately. This side-by-side comparison of real queries demonstrates how RAG augmentation improves faithfulness by reducing hallucination and retrieving corroborating evidence to support generative outputs.

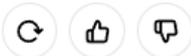

(a)

(b)

9.05  Spacer blocks shall be provided to ensure that the reinforcement is correctly positioned in the Works and shall be as small as possible consistent with their purpose. Such blocks shall not exceed 50 mm in length and their thickness shall be exactly that required to give cover specified on the Drawings. Each spacer block shall be made of concrete class 37.5/10 and binding wire shall be securely embedded into each block to permit adequate fastening to the reinforcement bars.   Precast Concrete

(c)

15. What is the formula for price adjustment for this contract?

GPT-4 Poe

The contract stipulates that it is NOT SUBJECT to price adjustment according to GCC Clause 44. This means there is no formula provided for price adjustment within this contract, and the contract price is fixed regardless of changes in cost or economic conditions.

(d)

15. What is the formula for price adjustment for this contract?

Pc = A c + B clmc/loc

(e)

| 44. Price Adjustment | 44.1 Prices shall be adjusted for fluctuations in the cost of inputs only if **provided for in the SCC**. If so provided, the amounts certified in each payment certificate, before deducting for Advance Payment, shall be adjusted by applying the respective price adjustment factor to the payment amounts due in each currency. A separate formula of the type indicated below applies to each Contract currency: |
|---|---|

$$P_c = A_c + B_c I_{mc}/I_{oc}$$

(f)

Figure 11: Screenshots of queries, responses, and original text from the contract document

### 6.3. Model limitation

While the retrieval-augmented GPT-4 model shows promising results in querying the construction contract document, some limitations need to be acknowledged. The model struggled to retrieve relevant passages for two of the questions, defaulting to uninformative "I don't know" non-answers. This indicates that the chunking strategies and semantic search techniques used were unable to adequately link some complex questions to supporting evidence in the contract document. Future studies can explore different embedding models and vector databases and compare their performance. Generalization is another limitation - the model was trained on just a single contract document and may fail to transfer to new projects with different terminology, formats, and content. Training on a large corpus of diverse contracts would likely improve out-of-domain robustness.

## 7. Conclusion

This research aimed to provide a comprehensive analysis of the current state, opportunities, and challenges of applying generative AI in the construction industry. Key insights were synthesized through a systematic literature review and expert Delphi study. The literature review revealed that generative AI adoption in construction is still in very early stages, with just a handful of initial studies exploring applications like information retrieval, project planning, hazard recognition, and risk assessment. However, the great potential of generative techniques like large language models was highlighted for enhancing productivity, accuracy, and automation across construction tasks.

The expert panel discussions further expanded on promising applications of generative AI in the construction industry during the pre-construction, construction, and post-construction phases. Opportunities were identified for major data types, including text, images, and video. At least seven potential opportunities for each data type were identified. For instance, the identified opportunities for image-text application include land survey data extraction, 3D model specification, blueprint digitization, daily progress image analysis, material quality assessment, inventory management, as-built documentation text extraction, warranty claim documentation, and visual inspection reports. The experts also outlined critical challenges that need to be addressed regarding domain knowledge, data, training, validation, integration, adoption, resources, and responsible governance.

A methodology was proposed to guide construction professionals in building customized generative AI solutions using their own proprietary data. The framework steps of data collection, curation, model development, evaluation, and deployment aim to make these powerful technologies more accessible for practical industry deployment. The value of the framework was demonstrated through a case study on applying generative models for enhanced querying of construction contract documents. The retrieval-augmented system (RAG) showed a significantly improved ability to extract accurate, relevant information from contracts through natural language queries compared to a baseline generative model (GPT-4). In terms of quality, relevance, and reproducibility, the RAG system outperforms the base GPT-4 model by 5.2, 9.4, and 4.8%, respectively.

While this study provides valuable insights into the application of generative AI in construction, certain limitations present opportunities for future work. The systematic literature review was confined to three databases - Scopus, Web of Science, and ScienceDirect - based on their structured interfaces and indexing criteria. Despite snowball searching, some relevant articles may have been missed by focusing on these sources. The expert Delphi panel size was also restricted due to resource constraints, although care was taken to recruit highly experienced professionals. For the case study, only a single base large language model and embedding technique were utilized due to API access costs. Testing multiple state-of-the-art models could reveal further performance gains. Future studies can build on these findings by expanding the literature review across more databases, recruiting larger expert panels as generative AI becomes more prominent in the construction industry, and experimenting with diverse generative architectures and embedding methods given sufficient computing power and budgets. Addressing these limitations represents an avenue for additional confirmatory research and comparative assessment on applying generative AI for construction tasks.

Nonetheless, this study provides a solid foundation of insights and a practical framework to guide further advancement.

## Acknowledgment

This research is supported by the Department of Building and Real Estate, The Hong Kong Polytechnic University, Hung Hom, and the Centre for Advances in Reliability and Safety (CAiRS), Hong Kong Science Park12/F, Building 19W, Pak Shek Kok, NT, Hong Kong, China.